\PassOptionsToPackage{hyperfootnotes=false}{hyperref}
\documentclass[a4paper,fleqn]{cas-dc}
\usepackage[numbers]{natbib}
\usepackage{graphicx}
\usepackage{xurl}
\usepackage{caption}
\usepackage{subcaption}
\usepackage{tikz}
\usepackage{amssymb}
\usepackage{amsmath}
\usepackage{booktabs}
\usepackage{float}
\usepackage{multirow}
\usepackage{makecell}
\usepackage{array}
\usepackage{colortbl}
\usepackage{xcolor}
\usepackage{soul}

\ExplSyntaxOn
\cs_set:Npn \__first_footerline:
{
  \group_begin:
  \small \sffamily
  \ifnum\theblind>0\relax\else
    \__short_authors: :~
  \fi
  {\rmfamily\itshape Preprint}
  \group_end:
}
\ExplSyntaxOff

\begin{document}
\def\floatpagepagefraction{1}
\def\textpagefraction{.001}

\let\WriteBookmarks\relax
\let\printorcid\relax
\shorttitle{SRUG for Medical Image Translation}
\shortauthors{Qiu et al.}

\title[mode=title]{SRUG: A Fusion-Driven Generator Network for Medical Image Translation}

\author[label1]{Xihe Qiu}
\author[label1]{Yang Dai}
\author[label1]{Xiaoyu Tan}
\author[label1]{Sijia Li}
\author[label2]{Fenghao Sun}
\cormark[3]
\author[label4,label5,label6]{Lu Gan}
\cormark[2]
\author[label3]{Liang Liu}
\cormark[1]

\cortext[corLL]{Corresponding author: 3111230017@fudan.edu.cn (L.Liu)}
\cortext[corLG]{Corresponding author:lugan@fudan.edu.cn (L.Gan)}
\cortext[corFS]{Corresponding author:Sun.fenghao@zs-hospital.sh.cn(F.Sun)}

\affiliation[label1]{organization={School of Electronic and Electrical Engineering, Shanghai University of Engineering Science},
            city={Shanghai},
            country={China}}
\affiliation[label2]{organization={Department of Thoracic Surgery, Zhongshan Hospital of Fudan University},
            city={Shanghai},
            postcode={200032},
            country={China}}
\affiliation[label3]{organization={Clinical Research Unit, Institute of Clinical Science, Zhongshan Hospital of Fudan University},
            city={Shanghai},
            postcode={200032},
            country={China}}
\affiliation[label4]{organization={Department of Medical Oncology, Zhongshan Hospital of Fudan University},
            city={Shanghai},
            postcode={200032},
            country={China}}
\affiliation[label5]{organization={Cancer center, Zhongshan Hospital of Fudan University},
            city={Shanghai},
            postcode={200032},
            country={China}}
\affiliation[label6]{organization={Fudan Zhangjiang Institute},
            city={Shanghai},
            postcode={201203},
            country={China}}
\begin{abstract}
MRI sequence synthesis aims to recover missing image contrast while preserving patient-specific anatomy. The choice of generation mechanism affects both optimization and the way source information reaches the synthesized image. In this study, we propose SRUG, a supervised standalone fusion-driven generator that learns a deterministic source-to-target mapping for paired MRI synthesis. Its direct reconstruction formulation removes generator--discriminator competition and requires neither variational latent sampling nor iterative diffusion denoising. To support structural fidelity within this formulation, SRUG uses a residual encoder adapted for image reconstruction, with a full-resolution convolutional stem, pooling-separated feature stages, and channel recalibration embedded in the residual branches. A nested multi-scale decoder (NMD) repeatedly fuses the resulting spatial features and predicts the target sequence through a single output head. L1 loss and Multi-Scale Structural Loss (MSS loss) jointly supervise the prediction, connecting the direct generation objective to anatomical detail recovery. Experiments on BraTS 2023 show competitive reconstruction fidelity and structural consistency across three MRI translation tasks. Architecture ablations assess the effects of channel recalibration, NMD, and structural supervision within the same generator framework. Additional evaluation on IXI supports applicability to another dataset and modality pair, while zero-shot testing on BraTS 2019 provides preliminary evidence of cross-dataset transferability. These results support direct supervised generation as a practical approach to paired MRI sequence synthesis, with the adapted encoding and reconstruction pathway providing the basis for structural preservation. The SRUG implementation is publicly available at https://github.com/RisingRich/SRUG.
\end{abstract}


\maketitle
\hypersetup{
  pdfsubject={SRUG preprint},
  pdfauthor={Xihe Qiu, Yang Dai, Xiaoyu Tan, Sijia Li, Fenghao Sun, Lu Gan, Liang Liu}
}

\section{Introduction}
Medical image-to-image translation aims to transform medical images from one domain to another while preserving structural integrity and diagnostic accuracy. This task plays a crucial role in clinical diagnosis, as it compensates for the lack of complete multimodal imaging data and enables physicians to make informed decisions based on limited imaging information.

Magnetic resonance imaging (MRI) is a widely used multimodal imaging technique for disease diagnosis and assessment. Different MRI sequences, such as T1-weighted, T2-weighted, proton density (PD), and fluid-attenuated inversion recovery (FLAIR), provide complementary contrast information that supports comprehensive clinical evaluation \cite{ref1, ref2, ref43}. When lesions appear simultaneously across multiple slices, 2.5D or 3D methods are often required for accurate diagnosis \cite{ref37}. Specifically, T1-weighted images provide useful anatomical contrast and can help characterize contrast enhancement and lesion morphology, while the conspicuity of blood products depends on hemorrhage stage and is often better supported by susceptibility-sensitive sequences such as T2*/GRE or SWI. T2-weighted images highlight pathological changes such as edema or inflammation \cite{ref1, ref5}. Proton density-weighted images (PDWI) mainly rely on the number of hydrogen protons within tissues, thereby reflecting proton distribution. Meanwhile, FLAIR sequences suppress cerebrospinal fluid signals, enhancing the visibility of lesions such as periventricular abnormalities, demyelinating diseases, or small vessel pathologies, making them particularly important in clinical neuroimaging. However, due to high scanning costs, time constraints, and patient safety concerns \cite{ref1, ref6}, it remains challenging in clinical practice to acquire multiple MRI modalities for the same patient. The lack of comprehensive multimodal information may adversely impact diagnostic accuracy and treatment planning. These limitations highlight the clinical demand for effective medical image translation techniques, which aim to synthesize missing MRI modalities from acquired ones, thereby enhancing diagnostic accuracy and treatment planning.

\begin{figure*}[pos=tbp]
    \includegraphics[width=\textwidth]{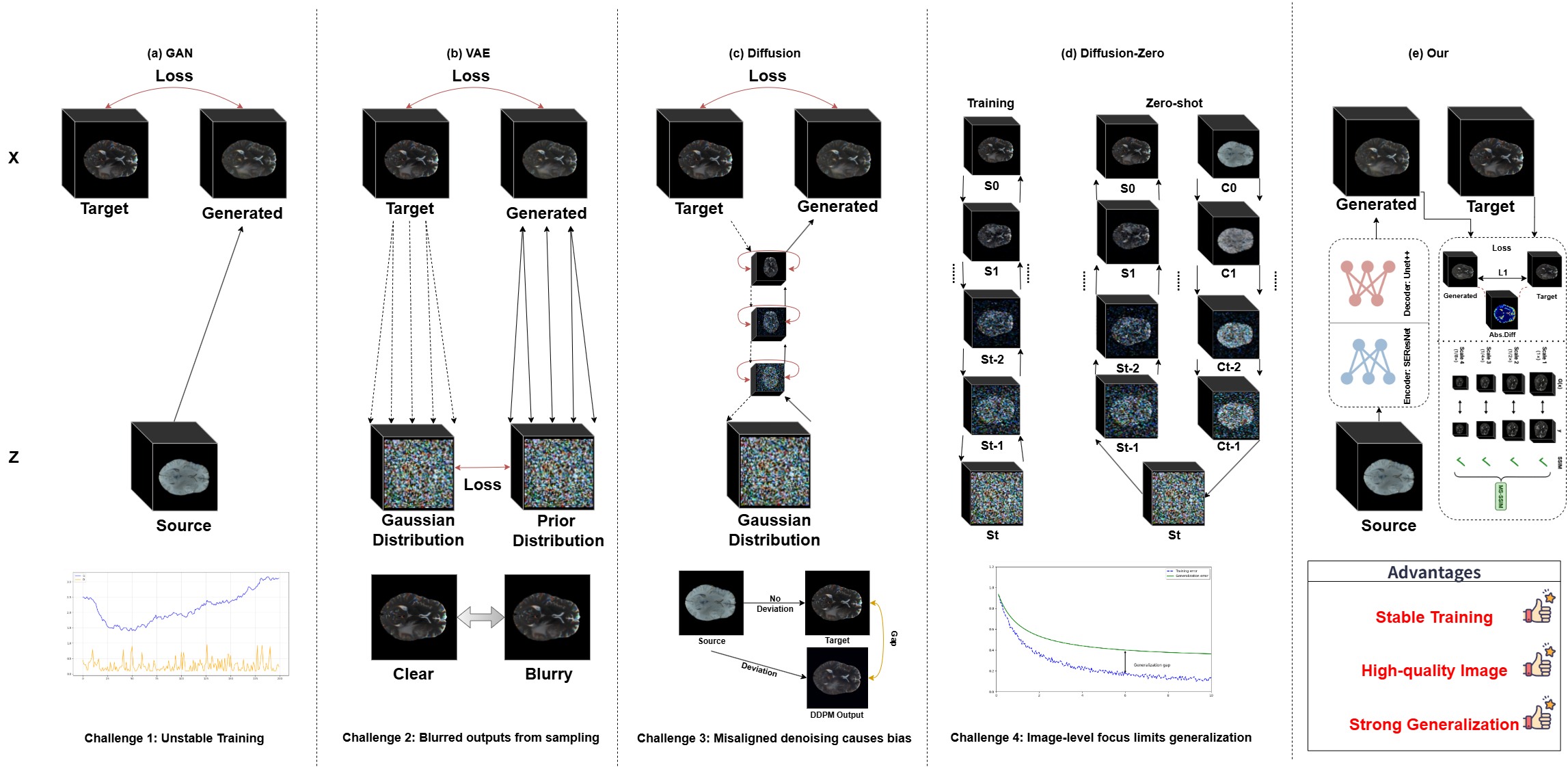}
\caption{Schematic motivation for SRUG. Panels (a)--(d) illustrate concerns associated with adversarial optimization, variational reconstruction, diffusion-based translation, and zero-shot transfer, respectively. Panel (e) presents direct source-to-target reconstruction using the CRRB-based encoder, NMD, and paired L1 and MSS loss supervision. The listed advantages express design aims; synthesis quality and transferability are evaluated in the experiments.}
\label{fig:0}
\end{figure*}

The generation mechanism is an important design choice for paired MRI translation (Fig.~\ref{fig:0}). GAN-based methods learn through coupled generator and discriminator objectives \cite{ref15, ref16, ref17}. This competition can complicate convergence and makes training sensitive to the balance between the two networks \cite{mescheder2018converge}. A directly supervised generator removes this source of optimization instability by learning against a fixed reconstruction criterion.

Variational models introduce a probabilistic latent representation and a prior-matching objective. Their reconstruction fidelity depends on the latent representation and the reconstruction criterion; element-wise reconstruction measures can yield overly smooth images with reduced perceptual detail \cite{larsen2016autoencoding}. This motivates retaining spatial source features throughout reconstruction and supervising structural similarity explicitly. Diffusion approaches instead construct an iterative denoising or bridge trajectory \cite{ref13, ref21, ref24}. Such trajectories support expressive generation but require a sampling procedure with repeated network evaluations. These mechanisms motivate a direct source-to-target alternative when aligned training pairs are available.

The resulting design question is whether a standalone generator can provide accurate MRI synthesis through direct supervised optimization while preserving the spatial detail needed for target-sequence reconstruction. SRUG addresses this question by combining a deterministic generation formulation with an encoder--decoder pathway that repeatedly reuses source-derived spatial features.

\begin{figure*}[pos=tbp]
    \includegraphics[width=\textwidth]{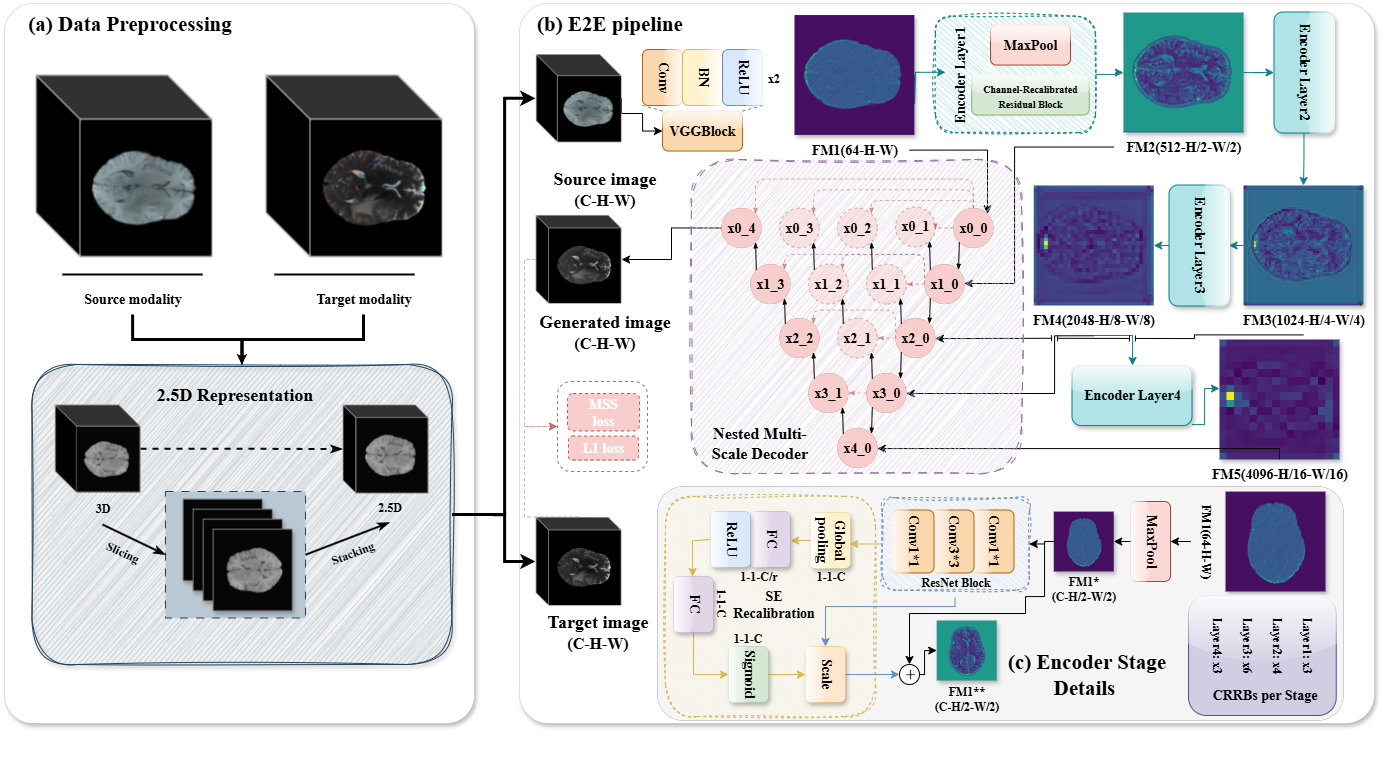}
    \caption{SRUG workflow: (a) 2.5D preprocessing; (b) the encoder built from channel-recalibrated residual blocks (CRRBs), reconstruction with the nested multi-scale decoder (NMD), and loss supervision; and (c) encoder stage details. MSS loss denotes Multi-Scale Structural Loss, computed from MS-SSIM.}
    \label{fig:1}
\end{figure*}

We propose SRUG, a supervised standalone generator for direct MRI sequence synthesis, as illustrated in Fig.~\ref{fig:1}. The model learns one deterministic mapping from the acquired sequence to its paired target. The reconstruction objective combines L1 loss with Multi-Scale Structural Loss (MSS loss), computed from MS-SSIM, allowing all generator parameters to be optimized without a competing discriminator. Inference consists of one forward evaluation, with no variational latent sampling or iterative denoising schedule.

The architecture is designed to support this direct reconstruction task. A residual encoder retains a full-resolution shallow feature and forms a spatial feature pyramid through pooling-separated stages. Channel recalibration operates inside the residual branches, so feature selection remains part of encoding. The nested multi-scale decoder (NMD) combines source-derived features with intermediate reconstructions through repeated fusion across the feature pyramid, allowing shallow detail and coarser context to contribute to the final prediction. Experiments on BraTS 2023 \cite{ref25}, independent training and testing on IXI \cite{ref27}, and zero-shot testing on BraTS 2019 \cite{ref5} evaluate the reconstruction quality and applicability of this direct generator.

The main contributions of this work can be summarized as follows:

\begin{itemize}
    \item We develop SRUG, a supervised standalone generator tailored to paired MRI sequence synthesis. It learns a direct source-to-target mapping with pixel-wise and structural supervision, removing adversarial optimization and producing a target image in one forward evaluation without variational sampling or an iterative diffusion trajectory.

    \item We adapt the encoding and reconstruction pathway to support anatomical detail recovery within this direct formulation. A full-resolution stem and pooling-separated residual stages supply spatial features, with channel recalibration embedded in the residual branches. NMD repeatedly reuses this feature pyramid and intermediate reconstructions through a single target-image output head.

    \item We evaluate SRUG across three BraTS 2023 translation tasks, an independent IXI modality-pair setting, and zero-shot transfer to BraTS 2019 using patient-level 2.5D inputs. Architecture and loss ablations examine which design choices support the reported synthesis quality.
\end{itemize}

\section{Related Work}

\subsection{Adversarial and Variational Medical Image Translation}
In traditional medical image translation tasks, generative adversarial networks (GANs) have achieved remarkable success by adversarially training a generator and a discriminator \cite{ref15, ref16, ref17}. The generator aims to synthesize visually realistic images, while the discriminator is trained to distinguish between real and generated ones. For instance, Pix2Pix \cite{ref16} exploits pixel-wise correspondence between paired images for synthesis, whereas CycleGAN \cite{ref17} overcomes the requirement of paired data and enables cross-modality translation even in unpaired scenarios. NICE-GAN \cite{ref14} shares the encoder structure between the generator and discriminator to improve efficiency and compactness. RegGAN \cite{ref18} integrates an image registration mechanism into the translation process to enhance anatomical consistency. ResViT \cite{ref19} incorporates vision transformer architectures to capture richer contextual features during translation. 
Recent image translation studies have also explored structure-preserving constraints, contrastive learning, Vision Mamba/state-space modeling, and self-cycle strategies for unpaired or cross-domain translation \cite{KANG2023109840, LIN2025111431, MA2026112253}.

Adversarial learning provides a learned signal for image realism, but the coupled optimization can exhibit convergence difficulties \cite{mescheder2018converge}. In paired translation, reconstruction losses provide an additional direct training signal. This distinction motivates examining how much synthesis quality can be obtained by optimizing the generator solely against the paired target, while adapting its architecture to preserve spatial information.

VAE-based methods and VAE--GAN hybrids have also been explored for MRI synthesis \cite{ref10, ref11}. A VAE couples reconstruction with regularization of a probabilistic latent representation. Work on learned reconstruction metrics shows that the choice of similarity measure substantially affects visual fidelity \cite{larsen2016autoencoding}. SRUG uses deterministic spatial features and explicit structural supervision, avoiding a variational sampling bottleneck while addressing detail preservation through its reconstruction pathway.

\subsection{Diffusion-Based Medical Image Translation}
Diffusion models learn expressive image distributions through denoising or score estimation \cite{ref13, ref20}. DDPMs generate images through successive reverse transitions, while DDIMs provide alternative sampling trajectories, including deterministic ones \cite{ref21}. In these iterative formulations, inference requires repeated denoiser evaluations, and the sampling schedule is part of the generation procedure.

SynDiff \cite{ref22} combines conditional diffusion with adversarial reverse projections to support large diffusion steps, and uses coupled diffusive and non-diffusive translation modules for learning from unpaired data. This approach illustrates how diffusion sampling and adversarial learning can be combined to improve medical image translation.

Fast-DDPM \cite{ref23} reduces the number of training and sampling time steps through an aligned scheduling strategy. BBDM \cite{ref24} models inter-domain translation using a Brownian bridge process, directly connecting the source and target domains. These developments improve the practicality of diffusion-based translation, while retaining a sequence of denoising or bridge updates.

SRUG reconstructs the target from paired MRI data through one encoder--decoder evaluation. Its residual feature pyramid and nested fusion pathway retain source-derived spatial information for direct reconstruction.

\section{Methods}
SRUG (Fig.~\ref{fig:1}) learns a deterministic mapping $G:x\mapsto\hat{y}$ from a source MRI slice stack to its target sequence. A full-resolution stem and four residual stages form a five-level feature pyramid, with SE recalibration embedded in the residual branches. NMD repeatedly fuses these features across resolutions to recover the target in one forward evaluation. L1 and MSS loss jointly supervise the prediction.

\subsection{Residual Encoding Adapted for Image Reconstruction}
The encoder uses channel-recalibrated residual blocks (CRRBs), combining residual bottlenecks \cite{ref35} with SE recalibration \cite{ref32}. Two $3\times3$ convolution--batch-normalization--ReLU layers form a 64-channel full-resolution stem. Four stages containing $[3,4,6,3]$ CRRBs follow, each preceded by $2\times2$ max pooling with stride two. All CRRB convolutions use stride one. This arrangement retains a full-resolution reference and places downsampling between residual stages, adapting the classification-oriented ResNet hierarchy for reconstruction.

Each CRRB contains a $1\times1$, $3\times3$, and $1\times1$ convolution sequence with an expansion factor of four. Batch normalization follows each convolution, and ReLU follows the first two convolutions. SE acts on the residual branch before shortcut addition:

\begin{equation}
h' = \operatorname{ReLU}\left(s(F(h))\odot F(h)+P(h)\right).
\end{equation}

Here, $F(h)$ is the residual transform, $\odot$ denotes channel-wise multiplication, and $P(h)$ is an identity shortcut or a $1\times1$ projection with batch normalization when channel dimensions differ. The gate $s$ uses global average pooling and two fully connected layers with reduction ratio 16, separated by ReLU and followed by sigmoid.

Let $X_{0,0}$ denote the stem feature and $E_l$ the residual stage at level $l$. The encoder pyramid is constructed as

\begin{equation}
\begin{aligned}
X_{0,0} &= H_{\mathrm{stem}}(x),\\
X_{l,0} &= E_l\left(\operatorname{MaxPool}_{2\times2}(X_{l-1,0})\right),\\
&\quad l=1,\ldots,4.
\end{aligned}
\end{equation}

The five encoder outputs have $[64,\allowbreak512,\allowbreak1024,\allowbreak2048,\allowbreak4096]$ channels at resolutions from $(H,W)$ to $(H/16,W/16)$. Each spatial feature map is supplied to the corresponding decoder level.

\subsection{Nested Multi-Scale Decoder (NMD)}
Inspired by the nested skip connections of U-Net++ \cite{ref34}, we construct a nested multi-scale decoder (NMD) over the encoder's feature pyramid to fuse multi-scale features for target MRI reconstruction. Nodes $X_{i,0}$ are supplied by the residual encoder, while nodes with $j>0$ perform reconstruction and feature fusion:

\begin{equation}
\begin{aligned}
X_{i,j} = {}&H_{i,j}\bigl([X_{i,0},X_{i,1},\ldots,\\
&\qquad X_{i,j-1},U(X_{i+1,j-1})]\bigr),\\
&0\leq i\leq3,\quad 1\leq j\leq4-i.
\end{aligned}
\end{equation}

Here, $[\cdot]$ denotes channel concatenation, $U$ is bilinear upsampling by a factor of two, and $H_{i,j}$ contains two $3\times3$ convolution--batch-normalization--ReLU layers. Fusion nodes have 64 channels at full resolution and 512, 1024, and 2048 channels at deeper levels. The nested connections repeatedly combine encoder features with previously fused features and deeper context.

The final prediction is obtained from the last full-resolution node:

\begin{equation}
\hat{y}=G(x)=\operatorname{Conv}_{1\times1}(X_{0,4}).
\end{equation}

The output projection maps 64 feature channels to the three target slice channels. Training supervises this single output head.

\subsection{Direct Supervision with Multi-Scale Structural Loss}
\label{sec:loss}
SRUG is trained using L1 loss and Multi-Scale Structural Loss (MSS loss), based on MS-SSIM \cite{ref29}. L1 constrains intensity correspondence, while MSS loss measures structural agreement across image resolutions. For predictions $\hat{y}=G(x)$ and targets $y$ in a batch of size $B$, with $C=3$ slice channels and spatial dimensions $H\times W$, the objective is
\begin{equation}
\begin{aligned}
\mathcal{L}_{\mathrm{total}}
&=\lambda_1\mathcal{L}_{L1}+\lambda_2\mathcal{L}_{\mathrm{MSS}},\\
\mathcal{L}_{L1}
&=\frac{1}{BCHW}\sum_{b,c,h,w}
\left|\hat{y}_{bchw}-y_{bchw}\right|.
\end{aligned}
\label{eq:training_objective}
\end{equation}
L1 compares the prediction with the target normalized to $[-1,1]$. MSS loss uses $u=(\hat{y}+1)/2$ and $v=(y+1)/2$, followed by repeated $2\times2$ average pooling to form $M=5$ image scales.

At scale $j$, let $\mu_u,\mu_v$, $\sigma_u^2,\sigma_v^2$, and $\sigma_{uv}$ denote local means, variances, and covariance computed using an $11\times11$ Gaussian window with standard deviation 1.5. The luminance and contrast--structure similarity maps are
\begin{equation}
\begin{aligned}
\ell_j&=\frac{2\mu_u\mu_v+C_1}{\mu_u^2+\mu_v^2+C_1},\\
q_j&=\frac{2\sigma_{uv}+C_2}{\sigma_u^2+\sigma_v^2+C_2},
\end{aligned}
\label{eq:mss_local_maps}
\end{equation}
where $C_1=(0.01)^2$ and $C_2=(0.03)^2$. For sample $b$ and channel $c$, let $\bar{q}_{j,bc}$ be the spatial mean of $q_j$ and $\bar{s}_{j,bc}$ the spatial mean of $\ell_j q_j$. Their multi-scale score is
\begin{equation}
S_{bc}=[\bar{s}_{M,bc}]_+^{w_M}
\prod_{j=1}^{M-1}[\bar{q}_{j,bc}]_+^{w_j},
\label{eq:mss_aggregation}
\end{equation}
where $[z]_+=\max(z,0)$ and the scale weights are $(0.0448,\allowbreak\,0.2856,\allowbreak\,0.3001,\allowbreak\,0.2363,\allowbreak\,0.1333)$. This combines contrast--structure agreement across scales with luminance agreement at the coarsest scale. The loss averages over samples and slice channels:
\begin{equation}
\begin{aligned}
\mathcal{L}_{\mathrm{MSS}}
&=1-\frac{1}{BC}\sum_{b=1}^{B}\sum_{c=1}^{C}S_{bc}\\
&=1-\operatorname{MS\text{-}SSIM}(u,v).
\end{aligned}
\label{eq:mss_loss}
\end{equation}
We use a $1:1$ ratio between the two loss weights and jointly optimize the encoder and decoder using the final reconstructed image.

\section{Experiment}
We evaluate reconstruction quality on three BraTS 2023 \cite{ref25} tasks (T1$\rightarrow$T2, T1$\rightarrow$FLAIR, and T2$\rightarrow$FLAIR), independently train and test on IXI \cite{ref27} for PD$\rightarrow$T2 translation, and test zero-shot transfer from BraTS 2023 to BraTS 2019 \cite{ref5} without fine-tuning.

All methods use the same patient-level splits, 2.5D inputs, and preprocessing. Training epochs, batch size, optimizer, and learning rate are kept consistent unless a baseline requires different settings. Official baseline implementations are used where available, with input and output channels adapted to the slice stacks.

\begin{table*}[pos=tbp]
\caption{Quantitative results across MRI translation settings. MSE is measured on [0,255].\label{tab:quantitative_all}}
\centering
\fontsize{8pt}{9.5pt}\selectfont
\setlength{\tabcolsep}{2.5pt}
\renewcommand{\arraystretch}{1.06}
\begin{tabular*}{\textwidth}{@{\extracolsep\fill}llcccccc@{\extracolsep\fill}}
\toprule
Task & Model & PSNR $\uparrow$ & SSIM $\uparrow$ & LPIPS $\downarrow$ & MS-SSIM $\uparrow$ & MSE $\downarrow$ & NMSE $\downarrow$ \\
\midrule
\multirow{6}{*}{\makecell[c]{\textbf{BraTS 2023}\\\textbf{T1$\rightarrow$T2}}} & CycleGAN & 18.9567$\pm$0.2842 & 0.6912$\pm$0.0093 & 0.1467$\pm$0.0058 & 0.6293$\pm$0.0286 & 861.8345$\pm$52.1906 & 0.4430$\pm$0.0278 \\
 & NICE-GAN & 23.8071$\pm$0.2556 & 0.8486$\pm$0.0069 & 0.1146$\pm$0.0017 & 0.8795$\pm$0.0043 & 290.8364$\pm$18.5864 & 0.1558$\pm$0.0116 \\
 & Pix2Pix & 25.1078$\pm$0.0320 & \underline{0.8711$\pm$0.0007} & 0.0885$\pm$0.0006 & 0.9018$\pm$0.0007 & 212.5699$\pm$2.0190 & \underline{0.1133$\pm$0.0014} \\
 & ResViT & 24.8802$\pm$0.1528 & 0.8660$\pm$0.0017 & 0.1419$\pm$0.0006 & 0.8992$\pm$0.0027 & 218.8124$\pm$7.9094 & 0.1157$\pm$0.0037 \\
 & BBDM & \underline{25.5541$\pm$0.0594} & 0.8272$\pm$0.0034 & \underline{0.0702$\pm$0.0002} & \underline{0.9240$\pm$0.0007} & \underline{200.5621$\pm$3.6000} & 0.1151$\pm$0.0033 \\
 & \textbf{Our} & \textbf{27.3122$\pm$0.0641} & \textbf{0.9161$\pm$0.0006} & \textbf{0.0519$\pm$0.0007} & \textbf{0.9436$\pm$0.0004} & \textbf{136.5155$\pm$2.1574} & \textbf{0.0737$\pm$0.0013} \\
\midrule
\multirow{6}{*}{\makecell[c]{\textbf{BraTS 2023}\\\textbf{T1$\rightarrow$FLAIR}}} & CycleGAN & 20.9014$\pm$0.5001 & 0.7856$\pm$0.0122 & 0.1095$\pm$0.0042 & 0.7981$\pm$0.0208 & 571.5507$\pm$61.3933 & 0.1570$\pm$0.0159 \\
 & NICE-GAN & 21.5745$\pm$0.1816 & 0.7872$\pm$0.0027 & 0.1133$\pm$0.0014 & 0.8197$\pm$0.0061 & 482.0957$\pm$23.5587 & 0.1331$\pm$0.0055 \\
 & Pix2Pix & 23.6115$\pm$0.0368 & 0.8518$\pm$0.0007 & 0.0911$\pm$0.0010 & 0.8832$\pm$0.0008 & 304.9670$\pm$3.3068 & 0.0842$\pm$0.0014 \\
 & ResViT & 23.3014$\pm$0.1033 & 0.8495$\pm$0.0020 & 0.1555$\pm$0.0014 & 0.8801$\pm$0.0020 & 321.2815$\pm$5.9920 & 0.0897$\pm$0.0011 \\
 & BBDM & \underline{25.1561$\pm$0.0678} & \underline{0.8812$\pm$0.0022} & \underline{0.0594$\pm$0.0003} & \underline{0.9300$\pm$0.0008} & \underline{220.3581$\pm$4.4482} & \underline{0.0608$\pm$0.0017} \\
 & \textbf{Our} & \textbf{25.3711$\pm$0.0657} & \textbf{0.8994$\pm$0.0009} & \textbf{0.0579$\pm$0.0010} & \textbf{0.9304$\pm$0.0007} & \textbf{212.2256$\pm$3.6206} & \textbf{0.0595$\pm$0.0020} \\
\midrule
\multirow{6}{*}{\makecell[c]{\textbf{BraTS 2023}\\\textbf{T2$\rightarrow$FLAIR}}} & CycleGAN & 21.1103$\pm$0.2555 & 0.7891$\pm$0.0051 & 0.1098$\pm$0.0023 & 0.8052$\pm$0.0132 & 564.7090$\pm$29.3443 & 0.1533$\pm$0.0082 \\
 & NICE-GAN & 22.1454$\pm$0.2654 & 0.8113$\pm$0.0025 & 0.1097$\pm$0.0040 & 0.8564$\pm$0.0066 & 444.2221$\pm$25.7541 & 0.1187$\pm$0.0070 \\
 & Pix2Pix & 24.2966$\pm$0.0820 & \underline{0.8734$\pm$0.0014} & 0.0818$\pm$0.0018 & 0.9027$\pm$0.0049 & 280.5355$\pm$5.8669 & 0.0766$\pm$0.0017 \\
 & ResViT & 23.8413$\pm$0.2275 & 0.8679$\pm$0.0028 & 0.1539$\pm$0.0012 & 0.9025$\pm$0.0032 & 294.8466$\pm$17.6315 & 0.0816$\pm$0.0044 \\
 & BBDM & \underline{25.7029$\pm$0.0218} & 0.7114$\pm$0.0027 & \underline{0.0551$\pm$0.0003} & \underline{0.9437$\pm$0.0003} & \underline{199.1350$\pm$1.6588} & \underline{0.0571$\pm$0.0008} \\
 & \textbf{Our} & \textbf{26.4500$\pm$0.1087} & \textbf{0.9231$\pm$0.0016} & \textbf{0.0453$\pm$0.0011} & \textbf{0.9504$\pm$0.0012} & \textbf{177.6753$\pm$6.6324} & \textbf{0.0499$\pm$0.0026} \\
\midrule
\multirow{6}{*}{\makecell[c]{\textbf{IXI}\\\textbf{PD$\rightarrow$T2}}} & CycleGAN & 28.6780$\pm$0.8157 & 0.9014$\pm$0.0118 & 0.0417$\pm$0.0064 & 0.9519$\pm$0.0078 & 96.8970$\pm$17.6520 & 0.0412$\pm$0.0074 \\
 & NICE-GAN & 26.0006$\pm$0.6243 & 0.8569$\pm$0.0111 & 0.0700$\pm$0.0023 & 0.9284$\pm$0.0063 & 172.7595$\pm$22.6739 & 0.0750$\pm$0.0100 \\
 & Pix2Pix & 28.9860$\pm$0.0802 & 0.9002$\pm$0.0014 & 0.0419$\pm$0.0005 & 0.9538$\pm$0.0008 & 87.7507$\pm$1.6320 & 0.0372$\pm$0.0008 \\
 & ResViT & 27.7821$\pm$0.0396 & 0.8966$\pm$0.0014 & 0.1067$\pm$0.0017 & 0.9461$\pm$0.0007 & 117.0431$\pm$0.9836 & 0.0500$\pm$0.0004 \\
 & BBDM & \underline{34.4619$\pm$0.0175} & \underline{0.9410$\pm$0.0010} & \underline{0.0128$\pm$0.0000} & \textbf{0.9863$\pm$0.0000} & \underline{26.5268$\pm$0.1003} & \textbf{0.0116$\pm$0.0000} \\
 & \textbf{Our} & \textbf{34.5718$\pm$0.1371} & \textbf{0.9585$\pm$0.0052} & \textbf{0.0124$\pm$0.0002} & \underline{0.9862$\pm$0.0005} & \textbf{26.3076$\pm$1.0040} & \underline{0.0118$\pm$0.0005} \\
\midrule
\multirow{6}{*}{\makecell[c]{\textbf{BraTS 2019}\\\textbf{T1$\rightarrow$T2}\\\textbf{zero-shot}}} & CycleGAN & 18.3283$\pm$0.4200 & 0.6932$\pm$0.0097 & 0.1575$\pm$0.0087 & 0.6334$\pm$0.0309 & 1032.0253$\pm$92.7939 & 0.3568$\pm$0.0312 \\
 & NICE-GAN & 21.8022$\pm$0.2634 & 0.8396$\pm$0.0050 & 0.1419$\pm$0.0035 & 0.8484$\pm$0.0063 & 560.1304$\pm$45.8699 & 0.1753$\pm$0.0134 \\
 & Pix2Pix & 22.1255$\pm$0.0225 & 0.8471$\pm$0.0009 & 0.1173$\pm$0.0007 & 0.8650$\pm$0.0004 & 504.8383$\pm$1.9846 & 0.1561$\pm$0.0005 \\
 & ResViT & 22.4429$\pm$0.1722 & \underline{0.8518$\pm$0.0029} & 0.1398$\pm$0.0025 & 0.8734$\pm$0.0037 & 474.3373$\pm$16.0190 & 0.1455$\pm$0.0052 \\
 & BBDM & \underline{23.2111$\pm$0.0076} & 0.8425$\pm$0.0013 & \textbf{0.0948$\pm$0.0001} & \underline{0.8904$\pm$0.0001} & \underline{435.6817$\pm$0.7392} & \underline{0.1304$\pm$0.0002} \\
 & \textbf{Our} & \textbf{23.3426$\pm$0.1221} & \textbf{0.8820$\pm$0.0021} & \underline{0.0954$\pm$0.0010} & \textbf{0.8945$\pm$0.0015} & \textbf{414.5736$\pm$9.4162} & \textbf{0.1232$\pm$0.0029} \\
\midrule
\multirow{6}{*}{\makecell[c]{\textbf{BraTS 2019}\\\textbf{T1$\rightarrow$FLAIR}\\\textbf{zero-shot}}} & CycleGAN & 19.1543$\pm$0.3269 & 0.7738$\pm$0.0052 & 0.1503$\pm$0.0020 & 0.7267$\pm$0.0140 & 885.1753$\pm$69.8652 & 0.3903$\pm$0.0417 \\
 & NICE-GAN & 19.8237$\pm$0.1463 & 0.7810$\pm$0.0010 & 0.1641$\pm$0.0014 & 0.7572$\pm$0.0057 & 751.1709$\pm$23.5775 & 0.3410$\pm$0.0125 \\
 & Pix2Pix & 20.2966$\pm$0.0203 & 0.7970$\pm$0.0002 & 0.1501$\pm$0.0004 & 0.7861$\pm$0.0013 & 679.2047$\pm$2.8066 & 0.3037$\pm$0.0038 \\
 & ResViT & 19.8760$\pm$0.0770 & \underline{0.7987$\pm$0.0009} & 0.1664$\pm$0.0005 & 0.7766$\pm$0.0022 & 734.7328$\pm$10.5215 & 0.3263$\pm$0.0045 \\
 & BBDM & \textbf{21.4323$\pm$0.0102} & 0.7958$\pm$0.0022 & \textbf{0.1242$\pm$0.0000} & \textbf{0.8285$\pm$0.0002} & \textbf{550.3436$\pm$0.6417} & \textbf{0.2446$\pm$0.0005} \\
 & \textbf{Our} & \underline{21.0411$\pm$0.0996} & \textbf{0.8326$\pm$0.0004} & \underline{0.1375$\pm$0.0011} & \underline{0.8199$\pm$0.0028} & \underline{587.9168$\pm$12.2326} & \underline{0.2589$\pm$0.0102} \\
\midrule
\multirow{6}{*}{\makecell[c]{\textbf{BraTS 2019}\\\textbf{T2$\rightarrow$FLAIR}\\\textbf{zero-shot}}} & CycleGAN & 19.9077$\pm$0.1378 & 0.7789$\pm$0.0026 & 0.1444$\pm$0.0022 & 0.7356$\pm$0.0073 & 772.0559$\pm$22.7204 & 0.3118$\pm$0.0167 \\
 & NICE-GAN & 18.7695$\pm$0.1819 & 0.7768$\pm$0.0017 & 0.1800$\pm$0.0032 & 0.7355$\pm$0.0058 & 1003.9244$\pm$37.4095 & 0.4313$\pm$0.0193 \\
 & Pix2Pix & 20.1334$\pm$0.0762 & 0.8083$\pm$0.0011 & 0.1419$\pm$0.0019 & 0.7829$\pm$0.0025 & 713.7919$\pm$14.6762 & 0.3099$\pm$0.0083 \\
 & ResViT & 20.2343$\pm$0.1398 & 0.8101$\pm$0.0011 & 0.1651$\pm$0.0019 & 0.7887$\pm$0.0046 & \underline{687.3273$\pm$17.7069} & 0.3083$\pm$0.0037 \\
 & BBDM & \underline{20.2559$\pm$0.0041} & \underline{0.8331$\pm$0.0001} & \textbf{0.1393$\pm$0.0000} & \underline{0.7910$\pm$0.0003} & 699.7294$\pm$0.9122 & \underline{0.3038$\pm$0.0008} \\
 & \textbf{Our} & \textbf{21.0160$\pm$0.0662} & \textbf{0.8374$\pm$0.0009} & \underline{0.1394$\pm$0.0007} & \textbf{0.8152$\pm$0.0020} & \textbf{593.6091$\pm$11.2667} & \textbf{0.2699$\pm$0.0123} \\
\bottomrule
\end{tabular*}
\vspace{0.2cm}
\end{table*}

\begin{figure*}[pos=tbp]
    \centering
    \includegraphics[width=\textwidth]{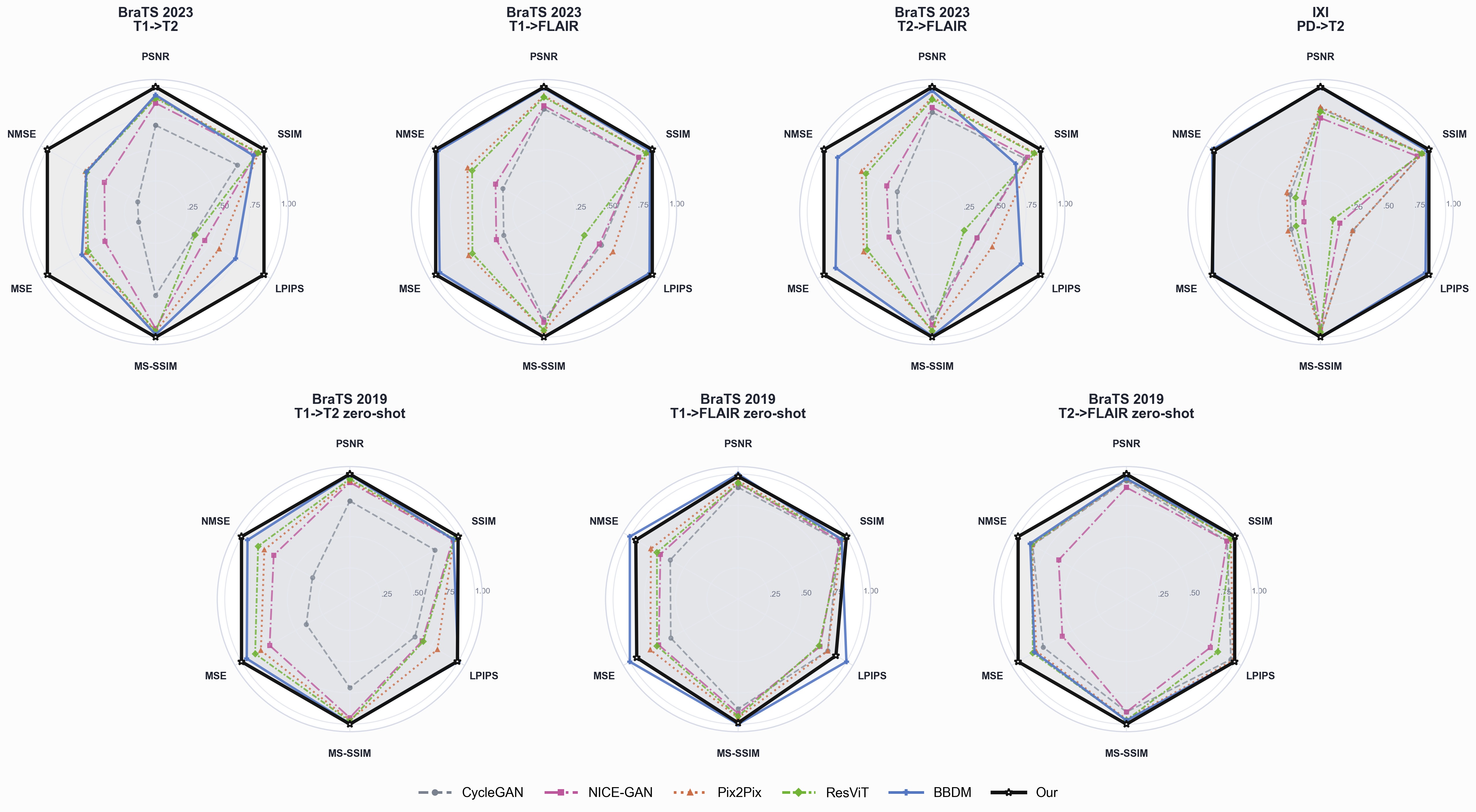}
    \caption{Normalized radar comparison across seven MRI translation settings and six metrics.}
    \label{fig:leida}
\end{figure*}

\subsection{Data Process}
For each patient and modality, the central axial slice and its two neighboring slices are extracted from the MRI volume and stacked as a three-channel 2.5D input \cite{ref37}. This provides local inter-slice context to the two-dimensional network. Each slice is independently normalized to [0,255] and resized to $256\times256$ using bilinear interpolation. Each patient contributes one source--target pair per translation task.

BraTS 2023 \cite{ref25} and IXI \cite{ref27} are split into non-overlapping patient groups for training and testing (Table~\ref{tab:data_splits}). Each BraTS 2023 task uses 237 training patients and 56 test patients. BraTS 2019 \cite{ref5} provides 335 cases per task for external zero-shot testing.

\begin{table*}[pos=tbp]
\centering
\caption{Patient-level train/test splits for MRI translation experiments.\label{tab:data_splits}}
\scriptsize
\setlength{\tabcolsep}{3pt}
\renewcommand{\arraystretch}{1.12}
\begin{tabular}{@{}p{0.16\textwidth}p{0.28\textwidth}>{\centering\arraybackslash}p{0.14\textwidth}>{\centering\arraybackslash}p{0.13\textwidth}p{0.21\textwidth}@{}}
\toprule
Dataset & Task & Train patients & Test patients & Split strategy \\
\midrule
\rowcolor{gray!6}
BraTS 2023 & \makecell[l]{T1$\rightarrow$T2\\T1$\rightarrow$FLAIR\\T2$\rightarrow$FLAIR} & 237 & 56 & patient-wise \\
IXI & PD$\rightarrow$T2 & 462 & 116 & patient-wise \\
\rowcolor{gray!6}
BraTS 2019 & \makecell[l]{T1$\rightarrow$T2\\T1$\rightarrow$FLAIR\\T2$\rightarrow$FLAIR} & -- & 335 & external testing \\
\bottomrule
\end{tabular}
\end{table*}

\subsection{Evaluation Metrics}
We evaluate generated images against paired targets using PSNR and SSIM \cite{ref28}, LPIPS \cite{ref30}, MS-SSIM \cite{ref29}, and MSE and NMSE \cite{ref31}. Results are reported as means and standard deviations on the independent test set. MSE is measured on the [0,255] intensity scale; the training-loss scales are defined in Section~\ref{sec:loss}.
\subsection{Experimental Setup}
Experiments were conducted on an NVIDIA GeForce RTX 3090 GPU.

Generator parameters were initialized with the Xavier method and optimized using Adam with learning rate $2\times10^{-4}$, $\beta_1=0.5$, and $\beta_2=0.999$. Models were trained for 200 epochs with batch size 2.

\section{Results}
To evaluate SRUG in a broad comparison setting, we selected representative baselines from several translation paradigms, including adversarial models such as CycleGAN \cite{ref17} and Pix2Pix \cite{ref16}, structurally enhanced architectures such as NICE-GAN \cite{ref14} and ResViT \cite{ref19}, and the diffusion-based BBDM \cite{ref24}. These baselines provide complementary reference points for assessing reconstruction fidelity, structural consistency, and robustness across MRI translation tasks.

\begin{figure*}[pos=tbp]
    \centering
    \includegraphics[width=0.95\textwidth]{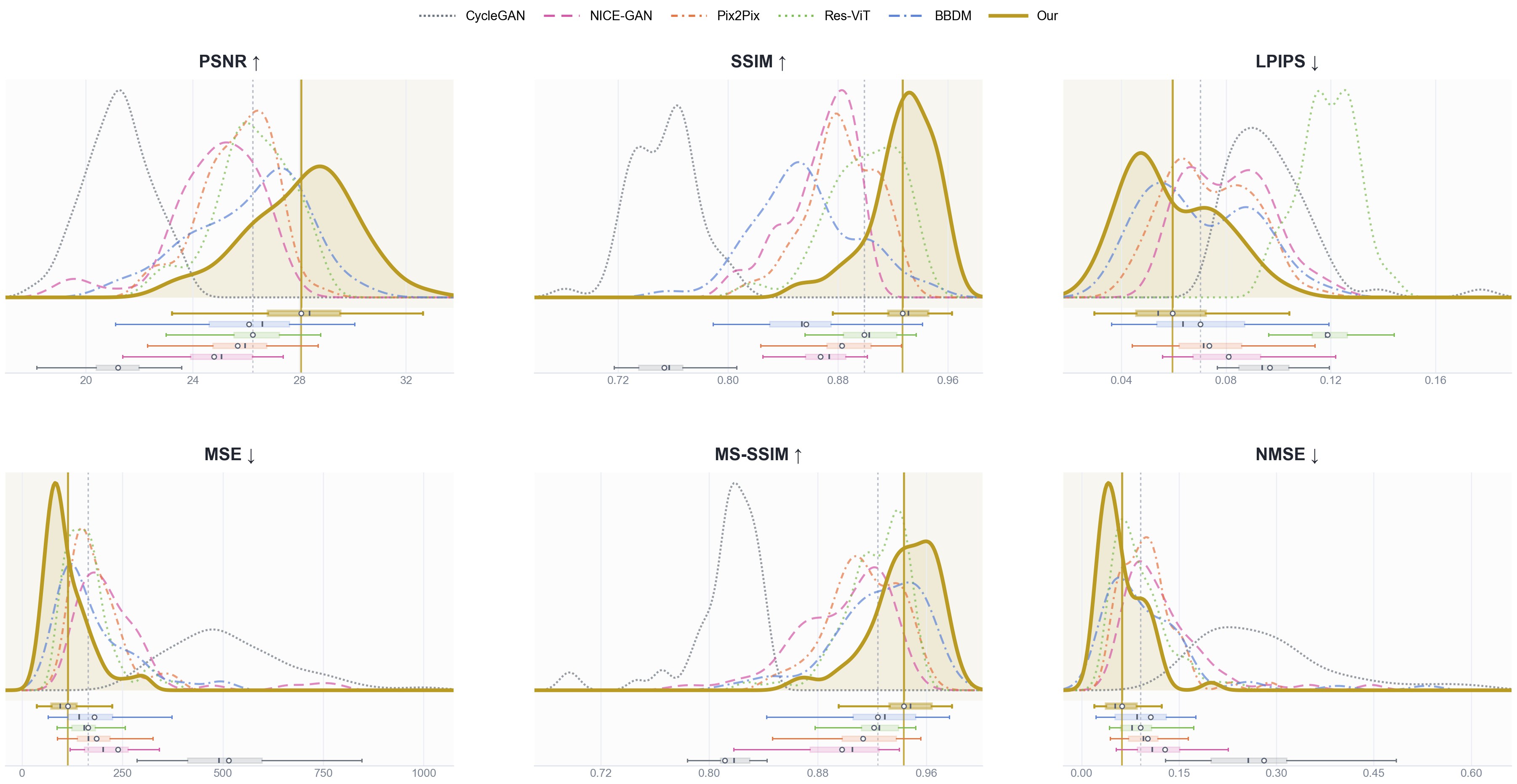}
    \caption{Metric distributions across MRI translation settings.}
    \label{fig:metric_distribution}
\end{figure*}

Figure~\ref{fig:metric_distribution} summarizes the metric distributions across methods and translation settings. This view complements Table~\ref{tab:quantitative_all} by showing metric dispersion rather than only mean values. Overall, SRUG tends to occupy favorable regions for structure-related metrics such as SSIM and MS-SSIM, while also maintaining competitive error-based performance for MSE and NMSE.

\begin{figure*}[pos=tbp]
    \centering
    \includegraphics[width=0.95\textwidth]{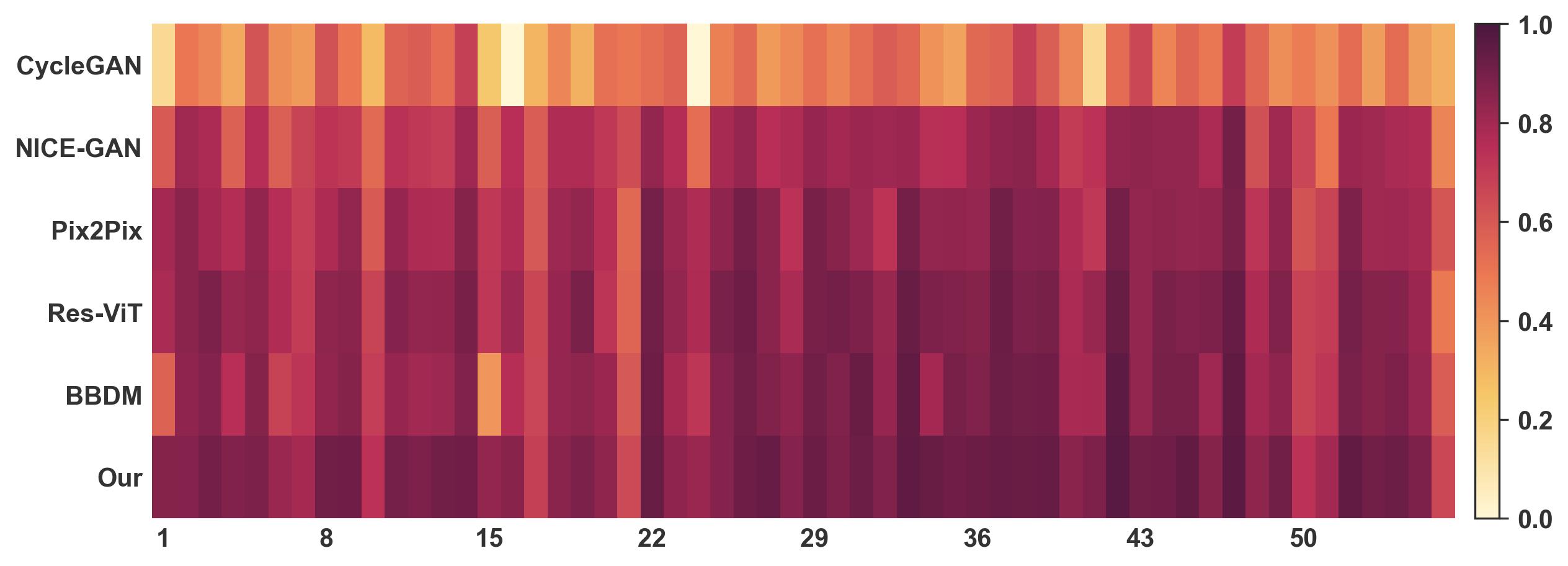}
    \caption{Pixel-intensity correlation with target images; darker colors indicate stronger correlation.}
    \label{fig:correlation_heatmap}
\end{figure*}

Figure~\ref{fig:correlation_heatmap} further evaluates the pixel-level intensity agreement between generated images and the Original target images. The Our row shows the highest overall correlation and appears darker across most samples, indicating that SRUG better preserves target-domain intensity patterns than the compared baseline methods.

\subsection{Multi-Task Evaluation on BraTS 2023}
On the BraTS 2023 \cite{ref25} dataset, we conducted systematic comparative experiments between the proposed method and multiple existing models (Table~\ref{tab:quantitative_all}). The experiments covered multi-task MRI translation (T1$\rightarrow$T2, T1$\rightarrow$FLAIR, T2$\rightarrow$FLAIR), each of which has clear clinical relevance: the different sequences are widely used in practice, and their registration is stable with well-aligned structures, providing a fair basis for evaluating model performance. The normalized radar charts in Fig.~\ref{fig:leida} provide a visual summary across all evaluated settings, while the main quantitative improvements are summarized below. Percentage changes are computed relative to the best baseline for each metric.

For T1$\rightarrow$T2 translation, SRUG achieved a PSNR of 27.3122, SSIM of 0.9161, LPIPS of 0.0519, MS-SSIM of 0.9436, MSE of 136.5155, and NMSE of 0.0737. Compared with the strongest baseline for each metric, SRUG improved PSNR, SSIM, and MS-SSIM by 6.88\%, 5.17\%, and 2.12\%, respectively, while reducing LPIPS, MSE, and NMSE by 26.07\%, 31.93\%, and 34.95\%. The paired significance analysis in Table~\ref{tab:significance_common} reports the task, strongest baseline, baseline value, SRUG value, absolute difference $\Delta$ (Our-base), improvement direction, Wilcoxon $p$ value, and Bonferroni-corrected $p$ value for each metric. For the T1$\rightarrow$T2 task, all Bonferroni-corrected $p$ values remained significant, supporting the statistical reliability of the observed improvements. Some metrics have identical or highly similar $p$ values because SRUG outperformed the strongest baseline for almost every paired case, resulting in identical or nearly identical extreme signed-rank statistics. The original paired metric values were checked, and these repeated $p$ values reflect the paired rank structure rather than a calculation or transcription error.

\begin{table*}[pos=tbp]
\centering
\caption{Wilcoxon significance tests for BraTS 2023 translation tasks.\label{tab:significance_common}}
\scriptsize
\setlength{\tabcolsep}{2pt}
\renewcommand{\arraystretch}{1.12}
\begin{tabular*}{\textwidth}{@{\extracolsep\fill}lllccclcc@{\extracolsep\fill}}
\toprule
Task & Metric & \makecell[c]{Strongest\\baseline} & \makecell[c]{Baseline\\value} & Our & $\Delta$ (Our-base) & Direction & Wilcoxon $p$ & Bonferroni $p$ \\
\midrule
\multirow{6}{*}{T1$\rightarrow$T2} & PSNR $\uparrow$ & BBDM & 25.5541 & 27.3122 & 1.7581 & Our higher & $8.459\times10^{-11}$ & $5.075\times10^{-10}$ \\
 & SSIM $\uparrow$ & Pix2Pix & 0.8711 & 0.9161 & 0.0450 & Our higher & $3.774\times10^{-11}$ & $2.264\times10^{-10}$ \\
 & LPIPS $\downarrow$ & BBDM & 0.0702 & 0.0519 & -0.0183 & Our lower & $3.774\times10^{-11}$ & $2.264\times10^{-10}$ \\
 & MS-SSIM $\uparrow$ & BBDM & 0.9240 & 0.9436 & 0.0196 & Our higher & $3.984\times10^{-11}$ & $2.391\times10^{-10}$ \\
 & MSE $\downarrow$ & BBDM & 200.5621 & 136.5155 & -64.0466 & Our lower & $3.984\times10^{-11}$ & $2.391\times10^{-10}$ \\
 & NMSE $\downarrow$ & Pix2Pix & 0.1133 & 0.0737 & -0.0396 & Our lower & $3.984\times10^{-11}$ & $2.391\times10^{-10}$ \\
\midrule
\multirow{6}{*}{T1$\rightarrow$FLAIR} & PSNR $\uparrow$ & BBDM & 25.1561 & 25.3711 & 0.2150 & Our higher & $8.805\times10^{-3}$ & $5.283\times10^{-2}$ \\
 & SSIM $\uparrow$ & BBDM & 0.8812 & 0.8994 & 0.0182 & Our higher & $2.555\times10^{-10}$ & $1.533\times10^{-9}$ \\
 & LPIPS $\downarrow$ & BBDM & 0.0594 & 0.0579 & -0.0015 & Our lower & $1.000\times10^{-4}$ & $6.000\times10^{-4}$ \\
 & MS-SSIM $\uparrow$ & BBDM & 0.9300 & 0.9304 & 0.0004 & Our higher & $3.357\times10^{-1}$ & $1.000$ \\
 & MSE $\downarrow$ & BBDM & 220.3581 & 212.2256 & -8.1325 & Our lower & $7.533\times10^{-3}$ & $4.520\times10^{-2}$ \\
 & NMSE $\downarrow$ & BBDM & 0.0608 & 0.0595 & -0.0013 & Our lower & $2.196\times10^{-2}$ & $1.318\times10^{-1}$ \\
\midrule
\multirow{6}{*}{T2$\rightarrow$FLAIR} & PSNR $\uparrow$ & BBDM & 25.7029 & 26.4500 & 0.7471 & Our higher & $2.415\times10^{-7}$ & $1.449\times10^{-6}$ \\
 & SSIM $\uparrow$ & Pix2Pix & 0.8734 & 0.9231 & 0.0497 & Our higher & $4.206\times10^{-11}$ & $2.524\times10^{-10}$ \\
 & LPIPS $\downarrow$ & BBDM & 0.0551 & 0.0453 & -0.0098 & Our lower & $5.221\times10^{-11}$ & $3.133\times10^{-10}$ \\
 & MS-SSIM $\uparrow$ & BBDM & 0.9437 & 0.9504 & 0.0067 & Our higher & $2.462\times10^{-6}$ & $1.477\times10^{-5}$ \\
 & MSE $\downarrow$ & BBDM & 199.1350 & 177.6753 & -21.4597 & Our lower & $6.150\times10^{-6}$ & $3.690\times10^{-5}$ \\
 & NMSE $\downarrow$ & BBDM & 0.0571 & 0.0499 & -0.0072 & Our lower & $4.061\times10^{-6}$ & $2.437\times10^{-5}$ \\
\bottomrule
\end{tabular*}
\end{table*}

For T1$\rightarrow$FLAIR translation, SRUG achieved a PSNR of 25.3711, SSIM of 0.8994, LPIPS of 0.0579, MS-SSIM of 0.9304, MSE of 212.2256, and NMSE of 0.0595. Relative to the strongest baseline for each metric, it improved PSNR, SSIM, and MS-SSIM by 0.85\%, 2.07\%, and 0.04\%, respectively, and reduced LPIPS, MSE, and NMSE by 2.53\%, 3.69\%, and 2.14\%. In the T1$\rightarrow$FLAIR part of Table~\ref{tab:significance_common}, the Bonferroni-corrected tests remained significant for SSIM, LPIPS, and MSE, were marginal for PSNR, and did not reach corrected significance for MS-SSIM or NMSE.

For T2$\rightarrow$FLAIR translation, SRUG achieved a PSNR of 26.4500, SSIM of 0.9231, LPIPS of 0.0453, MS-SSIM of 0.9504, MSE of 177.6753, and NMSE of 0.0499. Compared with the best baseline for each metric, SRUG improved PSNR, SSIM, and MS-SSIM by 2.91\%, 5.69\%, and 0.71\%, respectively, while reducing LPIPS, MSE, and NMSE by 17.79\%, 10.78\%, and 12.61\%. In the T2$\rightarrow$FLAIR part of Table~\ref{tab:significance_common}, all Bonferroni-corrected tests remained significant.

\begin{figure*}[pos=tbp]
    \centering
    \begin{subfigure}[b]{0.32\linewidth}
        \centering
        \includegraphics[width=\linewidth]{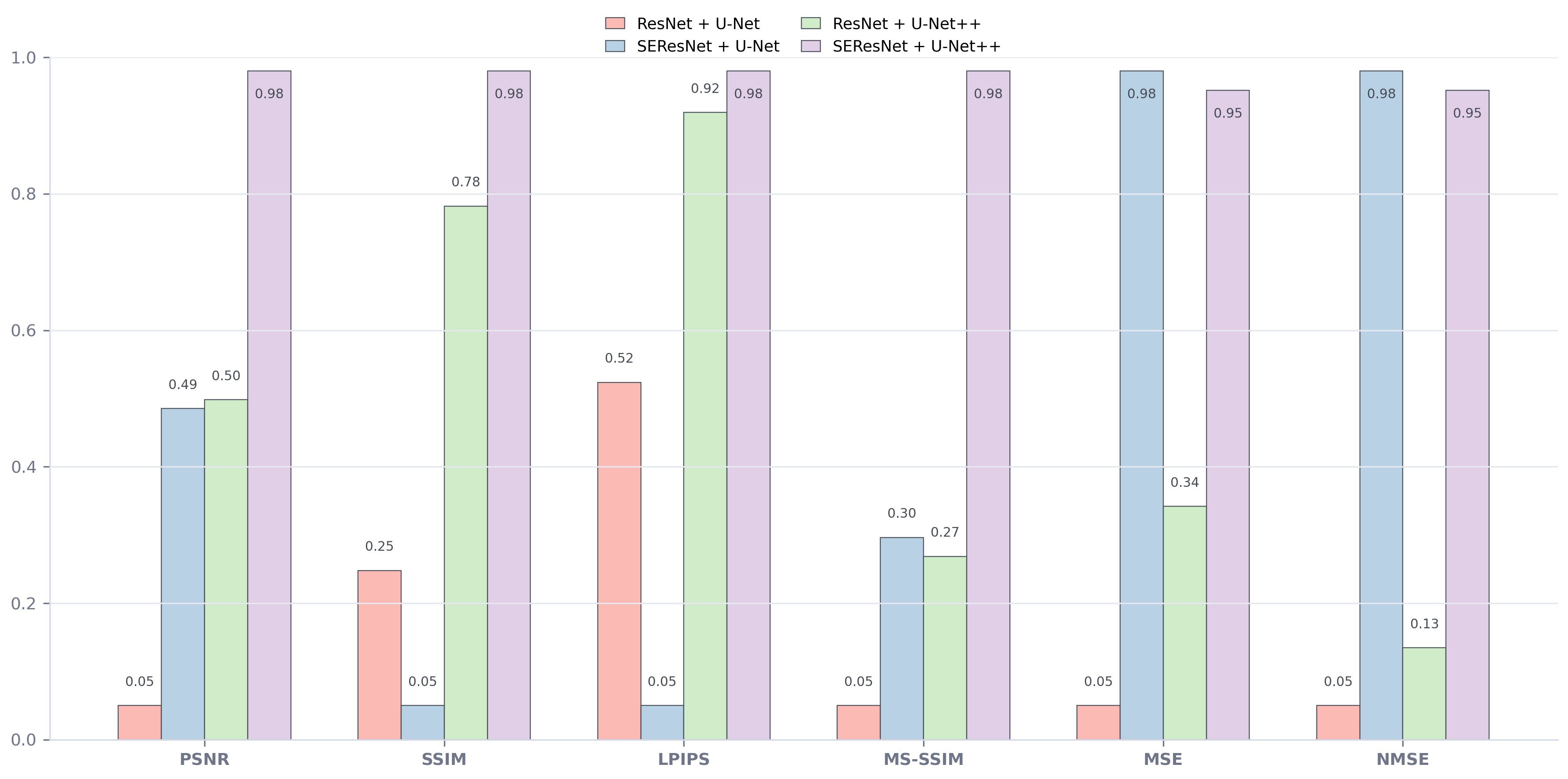}
        \caption{}
        \label{fig:bar_chart}
    \end{subfigure}
    \hfill
    \begin{subfigure}[b]{0.32\linewidth}
        \centering
        \includegraphics[width=\linewidth]{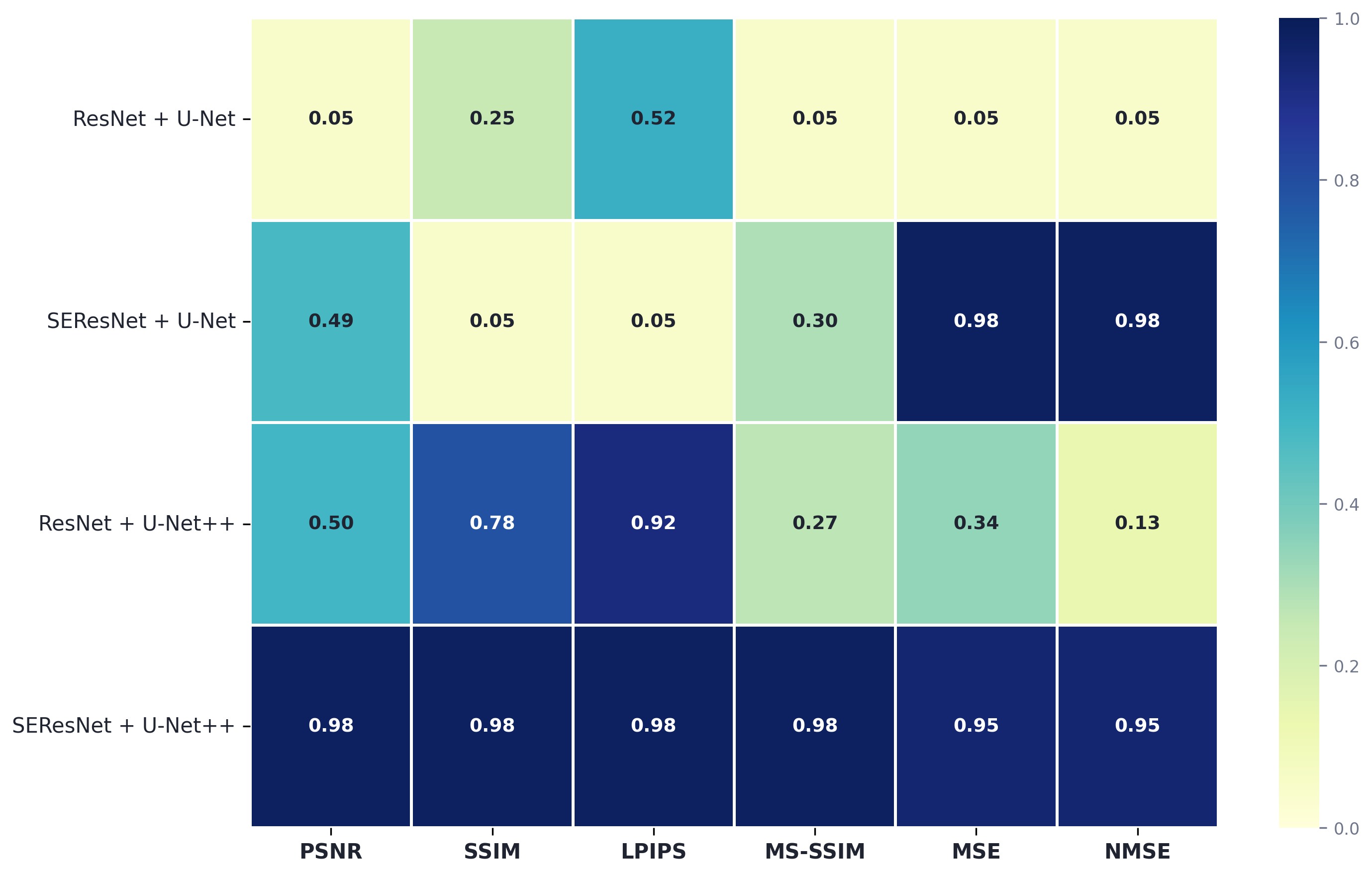}
        \caption{}
        \label{fig:heatmap}
    \end{subfigure}
    \hfill
    \begin{subfigure}[b]{0.32\linewidth}
        \centering
        \includegraphics[width=\linewidth]{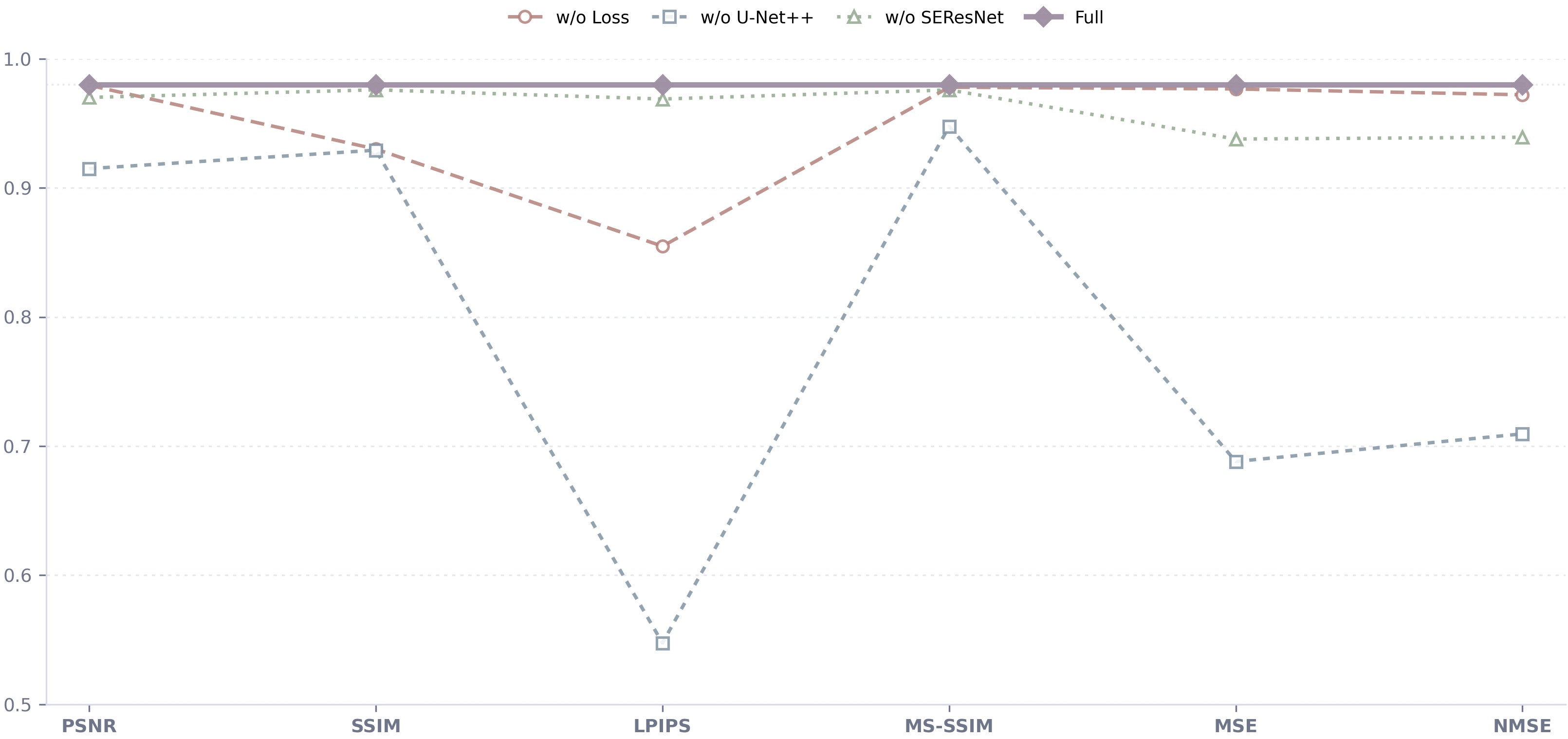}
        \caption{}
        \label{fig:component_ablation_profile}
    \end{subfigure}
    \caption{T1$\rightarrow$T2 ablation of channel recalibration, NMD, and MSS loss in SRUG: (a) normalized grouped bars; (b) normalized heatmap; (c) component-removal profiles. Configurations are defined in Table~\ref{tab:ablation}.}
    \label{fig:ablation}
\end{figure*}

\begin{figure*}[pos=tbp]
    \centering
    \begin{subfigure}[t]{0.9\textwidth}
        \includegraphics[width=\textwidth]{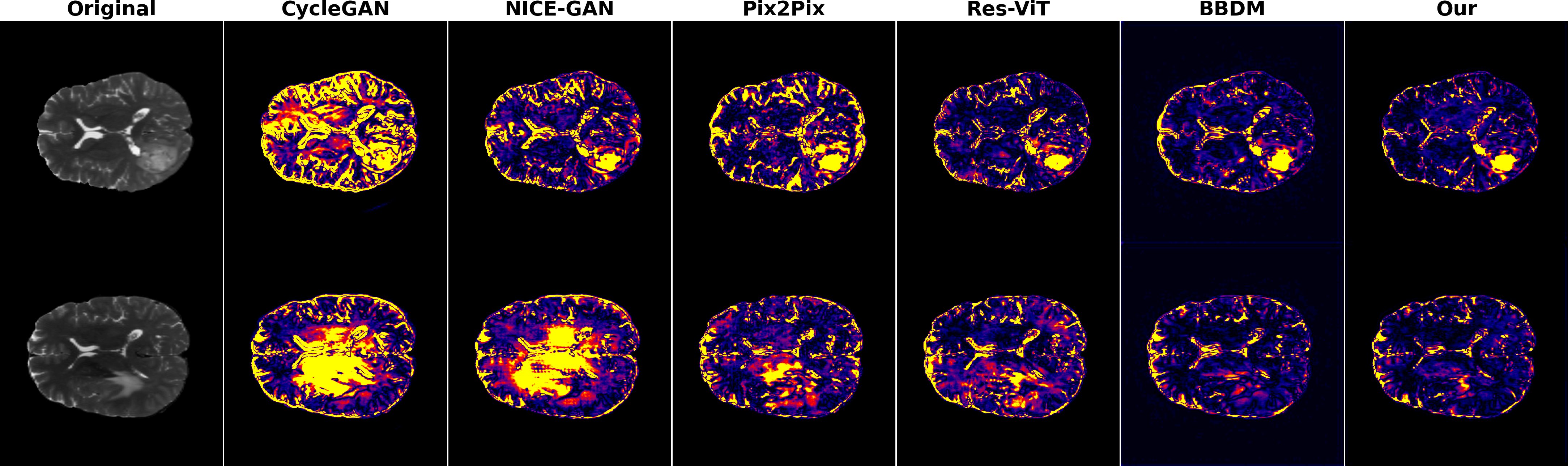}
    \end{subfigure}

    \begin{subfigure}[t]{0.9\textwidth}
        \includegraphics[width=\textwidth]{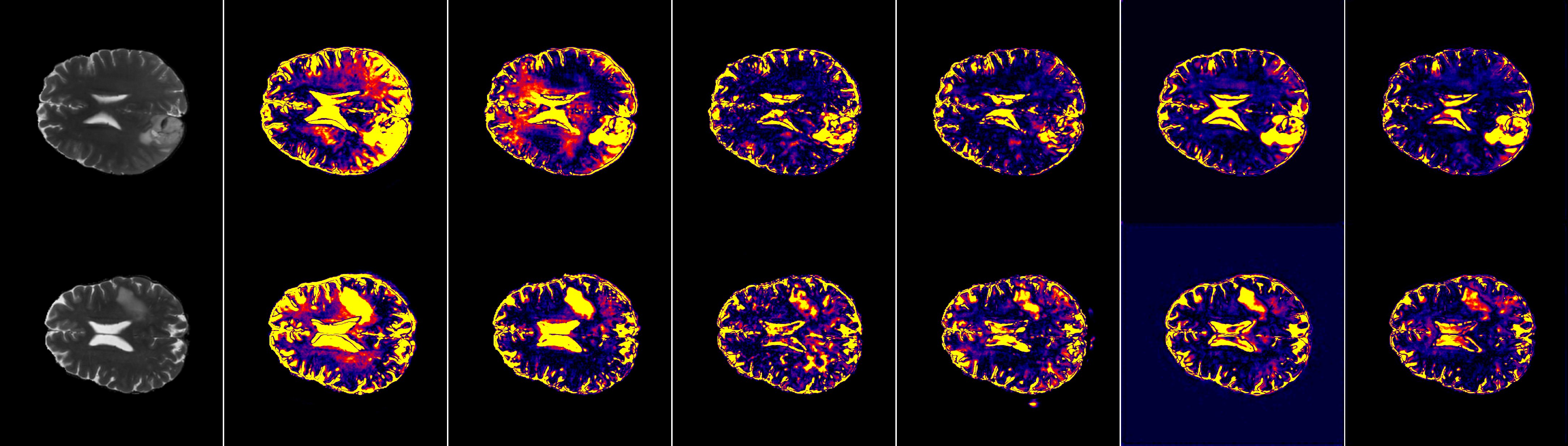}
    \end{subfigure}

    \begin{subfigure}[t]{0.9\textwidth}
        \includegraphics[width=\textwidth]{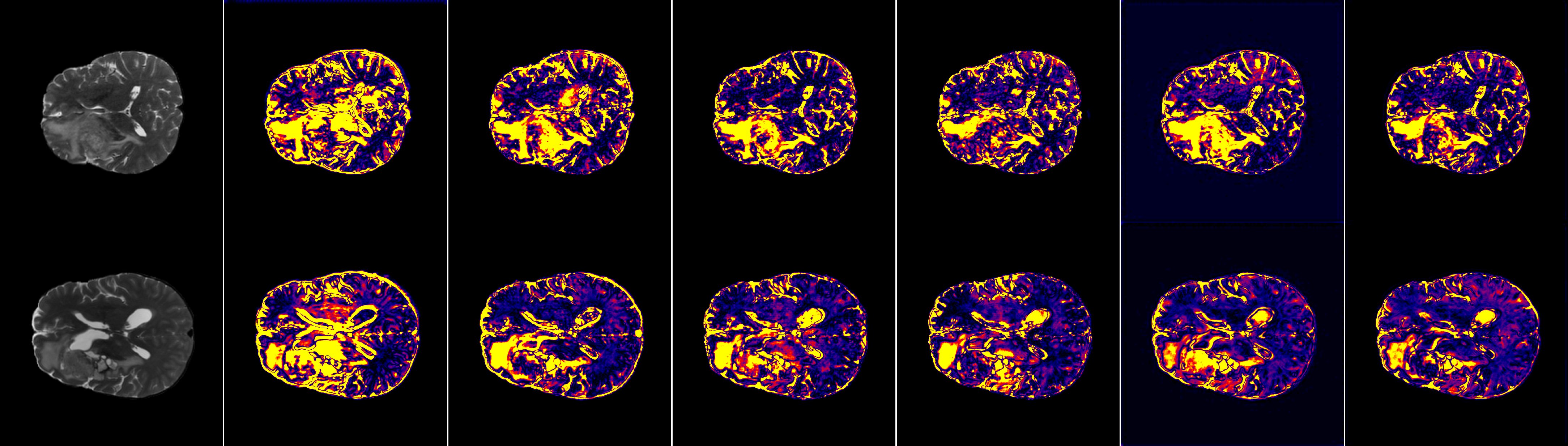}
    \end{subfigure}

    \begin{subfigure}[t]{0.9\textwidth}
        \includegraphics[width=\textwidth]{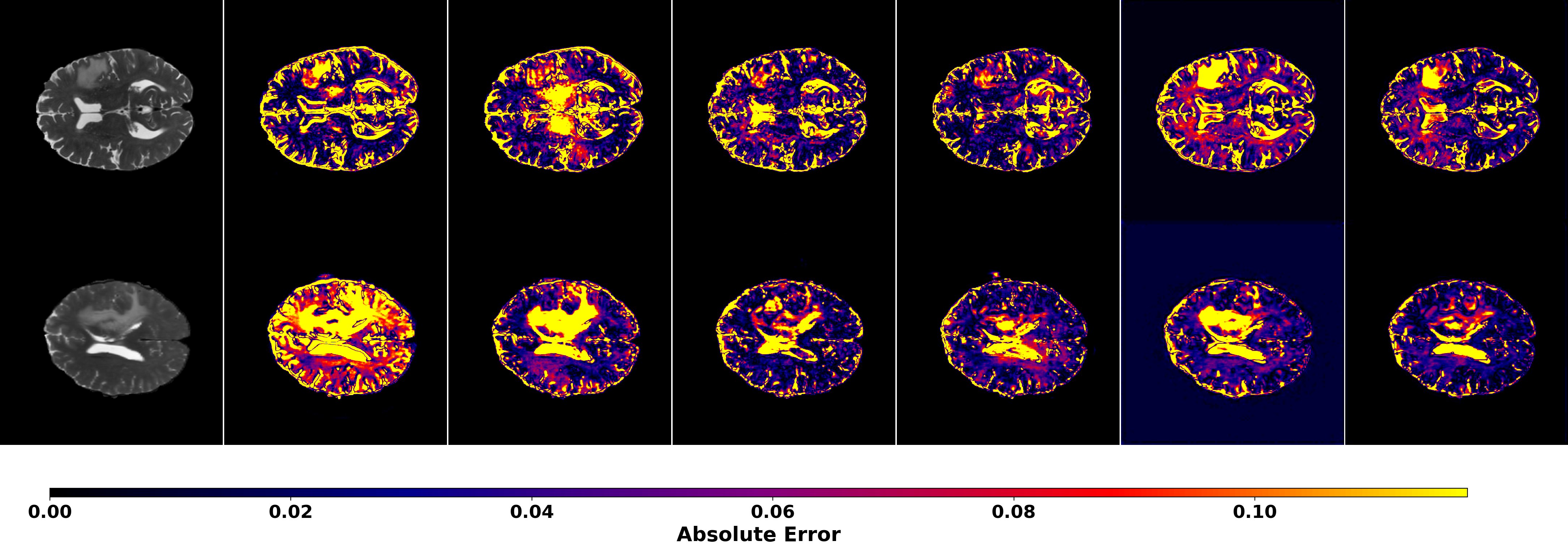}
    \end{subfigure}
    \caption{Full-image error heatmaps; brighter regions indicate larger errors.}
    \label{fig:heatmaps_part1}
\end{figure*}
\subsection{Additional Evaluation on IXI}

The proposed method was additionally evaluated on the IXI \cite{ref27} dataset for the PD$\rightarrow$T2 translation task (Table~\ref{tab:quantitative_all}). Models were trained and tested on IXI using the same patient-level split protocol. In this independent-dataset setting, SRUG achieved a PSNR of 34.5718, SSIM of 0.9585, LPIPS of 0.0124, MS-SSIM of 0.9862, MSE of 26.3076, and NMSE of 0.0118. Compared with the strongest baseline for each metric, SRUG improved PSNR and SSIM by 0.32\% and 1.86\%, respectively, and reduced LPIPS and MSE by 3.13\% and 0.83\%. BBDM retained small advantages on MS-SSIM and NMSE, with SRUG being 0.01\% lower in MS-SSIM and 1.72\% higher in NMSE. These results indicate that SRUG maintains strong structural similarity on IXI, while diffusion-based sampling remains competitive for selected structure and error metrics on this dataset.

\subsection{Zero-Shot Transfer on BraTS 2019}
To evaluate zero-shot cross-dataset transferability, we conducted testing on the BraTS 2019 \cite{ref5} dataset by directly applying models trained on BraTS 2023 \cite{ref25} without additional training or fine-tuning (Table~\ref{tab:quantitative_all}). On the T1$\rightarrow$T2 zero-shot test set with 335 cases, SRUG achieved a PSNR of 23.3426, SSIM of 0.8820, LPIPS of 0.0954, MS-SSIM of 0.8945, MSE of 414.5736, and NMSE of 0.1232. Compared with the best baseline for each metric, SRUG improved PSNR, SSIM, and MS-SSIM by 0.57\%, 3.55\%, and 0.46\%, respectively, and reduced MSE and NMSE by 4.84\% and 5.52\%. BBDM achieved a slightly lower LPIPS than SRUG in this setting, with SRUG's LPIPS being 0.63\% higher than the strongest baseline.

For the BraTS 2019 T1$\rightarrow$FLAIR zero-shot task, SRUG achieved a PSNR of 21.0411, SSIM of 0.8326, LPIPS of 0.1375, MS-SSIM of 0.8199, MSE of 587.9168, and NMSE of 0.2589. SRUG achieved the highest SSIM, improving over the strongest baseline by 4.24\%. BBDM achieved the best PSNR, LPIPS, MS-SSIM, MSE, and NMSE in this setting; relative to BBDM, SRUG was 1.83\% lower in PSNR, 1.04\% lower in MS-SSIM, 10.71\% higher in LPIPS, 6.83\% higher in MSE, and 5.85\% higher in NMSE.

For the BraTS 2019 T2$\rightarrow$FLAIR zero-shot task, SRUG achieved a PSNR of 21.0160, SSIM of 0.8374, LPIPS of 0.1394, MS-SSIM of 0.8152, MSE of 593.6091, and NMSE of 0.2699. Compared with the best baseline for each metric, SRUG improved PSNR, SSIM, and MS-SSIM by 3.75\%, 0.52\%, and 3.06\%, respectively, and reduced MSE and NMSE by 13.64\% and 11.16\%. BBDM achieved a slightly lower LPIPS than SRUG in this setting, with SRUG's LPIPS being 0.07\% higher than the strongest baseline.

\subsection{Experimental Analysis}
To further inspect local reconstruction behavior, we additionally visualized ROI-level error heatmaps, as shown in Fig.~\ref{fig:roi_heatmaps_part1}, with an additional ROI-level heatmap provided in Supplementary Fig.~\ref{fig:roi_heatmaps_part2}. Compared with full-image heatmaps, ROI-based visualization focuses on selected local anatomical regions and makes boundary-level deviations easier to observe. These local views are useful for assessing whether the model preserves fine structures in clinically relevant regions rather than only improving global image-level metrics.

\begin{figure*}[pos=tbp]
    \centering
    \includegraphics[width=\textwidth]{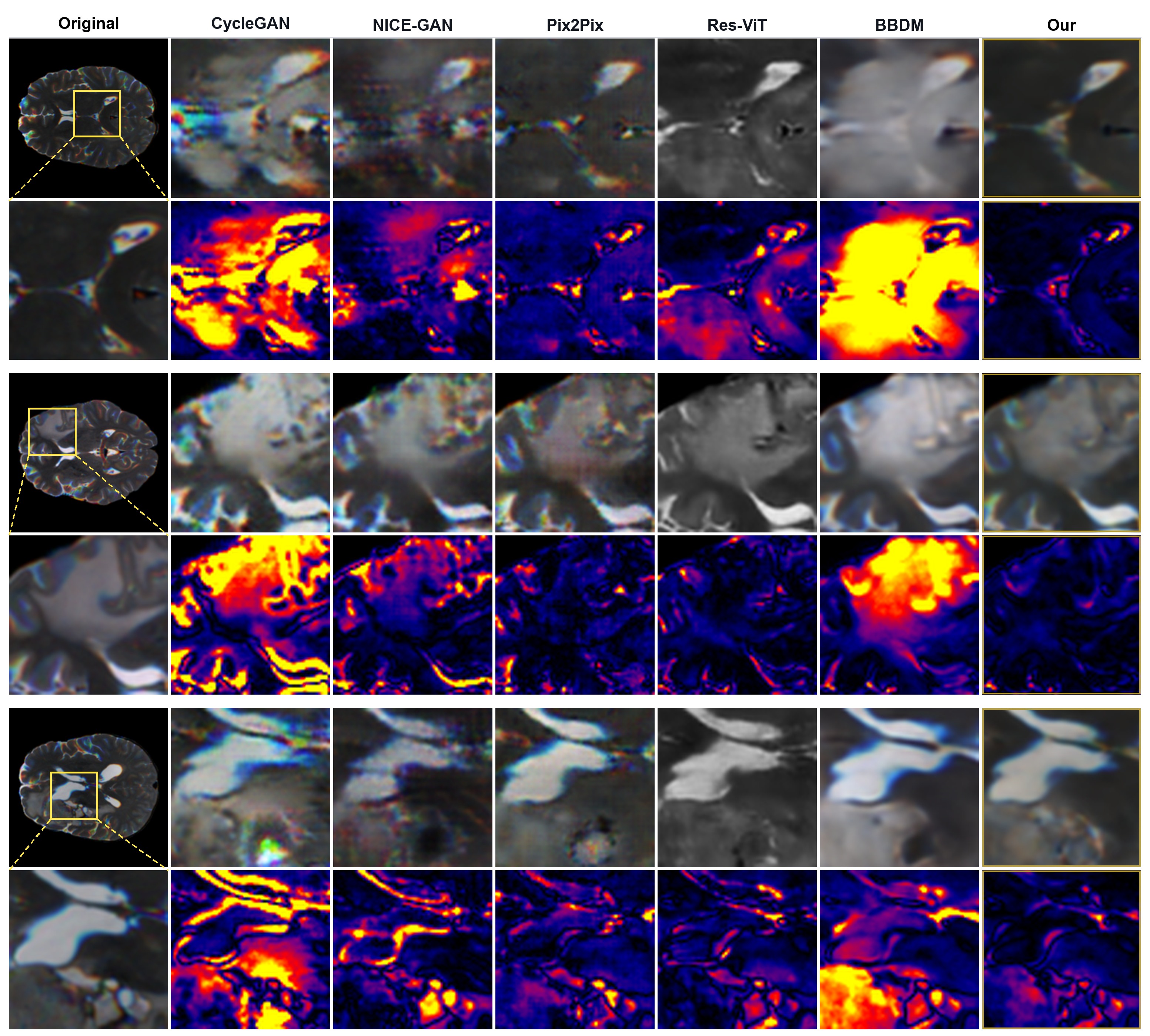}
    \caption{ROI-level error heatmaps in selected anatomical regions; brighter regions indicate larger errors.}
    \label{fig:roi_heatmaps_part1}
\end{figure*}

The ROI heatmaps show that error patterns are not uniformly distributed across the image, but tend to concentrate around local structural transitions and tissue boundaries. In these regions, SRUG produces relatively compact error responses, suggesting that the proposed fusion-driven generator better preserves local anatomical continuity. This observation is consistent with the improvements in SSIM and with the MS-SSIM gains observed in several evaluation settings, supporting the local reconstruction quality of the complete generator. The architecture ablations below examine the effects of its encoding and decoding choices.

As shown in Fig.~\ref{fig:heatmaps_part1}, we perform thermal error analysis on images generated by CycleGAN \cite{ref17} and Pix2Pix \cite{ref16} based on adversarial generation, NICE-GAN \cite{ref14} and ResViT \cite{ref19} based on structural improvements, BBDM \cite{ref24} based on the diffusion mechanism, as well as our proposed method, with the main experimental analysis conducted on this figure. To further demonstrate the consistency of these observations across different test cases, Fig.~\ref{fig:heatmaps_part2}–\ref{fig:heatmaps_part5} present supplementary thermal error visualizations on the test dataset, where each row corresponds to a representative case and each column to a different method.

\begin{table*}[pos=tbp]
\centering
\caption{T1$\rightarrow$T2 ablation of channel recalibration, NMD, and MSS loss in SRUG. Upper block: encoder--decoder variants; lower block: component removal. MSE is measured on [0,255].}
\label{tab:ablation}
\scriptsize
\setlength{\tabcolsep}{3pt}
\renewcommand{\arraystretch}{1.12}
\begin{tabular*}{\textwidth}{@{\extracolsep\fill}cccc|cccccc@{\extracolsep\fill}}
\toprule
\multicolumn{4}{c|}{\textbf{Generator Configuration}} & \multicolumn{6}{c}{\textbf{Image Quality Assessment}} \\
\cmidrule(lr){1-4}\cmidrule(lr){5-10}
\makecell[c]{Encoder\\w/o SE} & \makecell[c]{CRRB\\encoder} & \makecell[c]{Plain\\decoder} & NMD & PSNR $\uparrow$ & SSIM $\uparrow$ & LPIPS $\downarrow$ & MS-SSIM $\uparrow$ & MSE $\downarrow$ & NMSE $\downarrow$ \\
\midrule
\multicolumn{10}{l}{\textit{Replacement experiments: module substitution under the same generator pipeline.}} \\
$\checkmark$ &  & $\checkmark$ &  & 26.9974 & 0.9050 & 0.0624 & 0.9402 & 142.7445 & 0.0769 \\
 & $\checkmark$ & $\checkmark$ &  & 27.1447 & 0.9020 & 0.0733 & \underline{0.9411} & \textbf{136.3203} & \textbf{0.0736} \\
$\checkmark$ &  &  & $\checkmark$ & \underline{27.1491} & \underline{0.9131} & \underline{0.0533} & 0.9410 & 140.7279 & 0.0766 \\
 & $\checkmark$ &  & $\checkmark$ & \textbf{27.3122} & \textbf{0.9161} & \textbf{0.0519} & \textbf{0.9436} & \underline{136.5155} & \underline{0.0737} \\
\midrule
\multicolumn{10}{l}{\textit{Component-removal ablations; the full SRUG reference is shown in the final row above.}} \\
\multicolumn{4}{l|}{\textit{w/o MSS loss}} & 27.2960 & 0.8694 & 0.0595 & 0.9416 & 136.9923 & 0.0743 \\
\multicolumn{4}{l|}{\textit{w/o NMD}} & 25.4981 & 0.8686 & 0.0929 & 0.9124 & 194.4220 & 0.1018 \\
\multicolumn{4}{l|}{\textit{w/o CRRB encoder}} & 27.0339 & 0.9123 & 0.0525 & 0.9396 & 142.6604 & 0.0769 \\
\bottomrule
\end{tabular*}
\end{table*}

This visualization highlights pixel-wise discrepancies between generated outputs and the ground truth, where darker regions correspond to smaller errors and brighter regions indicate larger errors. Unsupervised adversarial generation methods, such as CycleGAN \cite{ref17} and NICE-GAN \cite{ref14}, tend to produce unstable local reconstructions and may fail to preserve fine anatomical boundaries. Supervised adversarial methods, such as Pix2Pix \cite{ref16}, benefit from paired training data and recover more local structure, but can still produce imprecise edge transitions. Hybrid structural models, such as ResViT \cite{ref19}, further improve boundary reconstruction, although visible gaps remain in some regions compared with the ground truth.

BBDM \cite{ref24} produces competitive image reconstructions and achieves the best values for several metrics in the external evaluation settings. Its relative performance varies across tasks and metrics, as shown in Table~\ref{tab:quantitative_all}. Differences in the displayed error maps describe reconstruction outcomes; they do not by themselves identify a particular sampling mechanism as the cause of local or background errors.

SRUG obtains its predictions through direct supervised reconstruction. Its full-resolution feature reference and nested reuse of the encoder pyramid provide spatial information for recovering boundaries and local contrast, while channel recalibration changes the responses available to these fusion operations. The observed reconstruction results show that this generator can preserve image structure without adversarial supervision or iterative denoising. The following ablations examine the contribution of its architecture and structural objective within the same direct generation framework.

\section{Ablation Study}
To analyze how the SRUG encoding and reconstruction pathway contributes to performance, we conducted an ablation study on the BraTS 2023 \cite{ref25} dataset using the T1$\rightarrow$T2 translation task. Table~\ref{tab:ablation} contains two complementary experimental blocks. The first block compares adapted residual encoders without and with SE, and a plain skip-connected decoder with NMD. Both residual encoder variants retain the same stem, stage widths, pooling schedule, and block counts. These comparisons assess channel recalibration and decoder connectivity within the SRUG architecture. The second block removes MSS loss, replaces NMD with a no-skip decoder, or replaces the SE-residual encoder with a TinyCNN. In Fig.~\ref{fig:ablation}, ResNet/SEResNet identify the encoder without channel recalibration and the CRRB-based encoder, respectively; U-Net/U-Net++ identify the plain decoder and NMD. The w/o U-Net++ label denotes removal of NMD. The results are also visualized in Fig.~\ref{fig:ablation}, which includes a normalized grouped bar chart (Fig.~\ref{fig:bar_chart}), a heatmap (Fig.~\ref{fig:heatmap}), and a reference-profile line plot (Fig.~\ref{fig:component_ablation_profile}).

All replacement variants share the direct supervised generation formulation, allowing the comparison to focus on encoding and decoder connectivity. The adapted residual encoder with the plain decoder achieves a PSNR of 26.9974, SSIM of 0.9050, and LPIPS of 0.0624. Enabling SE with this decoder improves PSNR, MS-SSIM, MSE, and NMSE, but reduces SSIM to 0.9020 and increases LPIPS to 0.0733. With the residual encoder unchanged, NMD improves SSIM to 0.9131 and LPIPS to 0.0533. Adding SE with NMD further improves these metrics to 0.9161 and 0.0519. Thus, the observed structural and perceptual benefit of recalibration depends on the decoding configuration. The full SRUG configuration achieves the highest PSNR and MS-SSIM values of 27.3122 and 0.9436, while its MSE and NMSE of 136.5155 and 0.0737 remain close to the best values obtained with the plain decoder. These results support the selected encoding--decoding configuration for this task; they do not establish a statistically significant interaction between the two design choices.

In the component-removal ablation, the w/o NMD setting reconstructs the output from the deepest encoder feature $x_{4,0}$ through four consecutive upsampling stages, each followed by a two-convolution fusion block. This removes both nested fusion and access to encoder skip features; the replacement comparison between the plain decoder and NMD above provides a complementary connectivity comparison. The w/o CRRB encoder setting, labeled w/o SEResNet in the plot, substitutes a lightweight TinyCNN for the entire encoder. This comparison changes the encoder family and capacity, while the residual-versus-SE-residual replacement above assesses the specific effect of channel recalibration. The w/o MSS loss setting, labeled w/o Loss in the plot, retains the generator and L1 supervision while setting $\lambda_2=0$ in Eq.~\ref{eq:training_objective}. Under these settings, w/o NMD shows the largest degradation, with PSNR decreasing to 25.4981 and MSE increasing to 194.4220. The TinyCNN encoder variant remains closer to the full model but still reduces PSNR, SSIM, and MS-SSIM to 27.0339, 0.9123, and 0.9396, respectively. Removing MSS loss mainly affects structural similarity, reducing SSIM from 0.9161 to 0.8694. These observations support the contributions of the adapted encoder, nested reconstruction pathway, and structural supervision to the reported synthesis results.

\subsection{Ablation Study Analysis}
\begin{figure*}[pos=tbp]
    \centering
    \includegraphics[width=\textwidth]{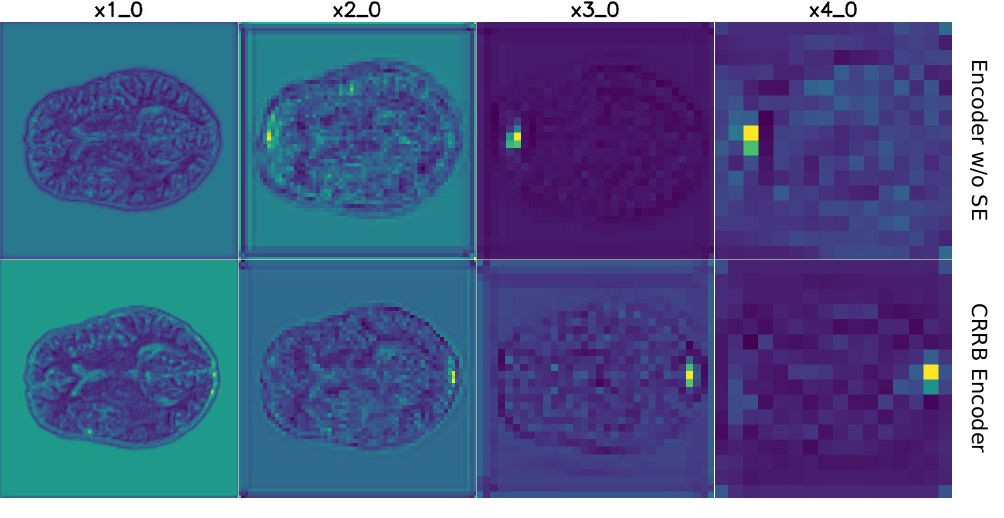}
    \caption{Effect of channel recalibration on encoder feature maps at levels $x_{1,0}$--$x_{4,0}$. Top: encoder variant without channel recalibration. Bottom: encoder built from CRRBs.}
    \label{fig:heatmaps_stage0}
\end{figure*}

\begin{figure*}[pos=tbp]
    \centering
    \includegraphics[width=\textwidth]{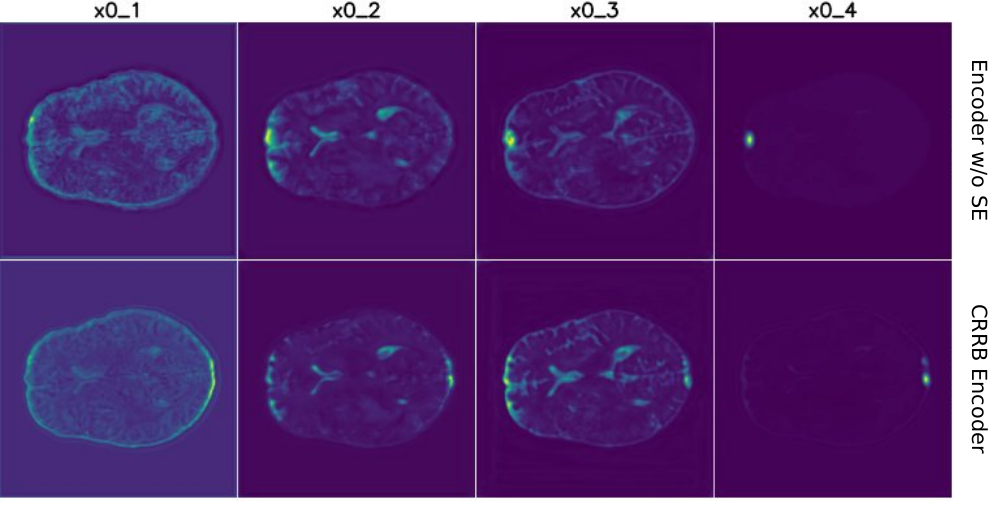}
    \caption{Effect of encoder channel recalibration on NMD feature maps at nodes $x_{0,1}$--$x_{0,4}$. Top: features from the encoder variant without channel recalibration. Bottom: features from the encoder built from CRRBs.}
    \label{fig:heatmaps_decoder}
\end{figure*}

Based on the ablation experiments, NMD provides an effective reconstruction pathway, while channel recalibration changes the feature responses supplied to that pathway. Fig.~\ref{fig:heatmaps_stage0} and Fig.~\ref{fig:heatmaps_decoder} compare the adapted residual encoder without and with SE at different encoding and decoding nodes. The encoder hierarchy and NMD are shared across these variants, providing a visual comparison of how residual-branch recalibration affects the translation process.

In the encoding stage, the shallow features (x1\_0, x2\_0) extracted by the SE-residual encoder preserve finer details and exhibit more discriminative feature responses compared with the residual encoder without SE. At the middle level (x3\_0), the encoder without SE tends to produce broader and more ambiguous activations, whereas the SE-residual encoder generates clearer intermediate representations. At the deepest layer (x4\_0), the SE-residual encoder shows a more concentrated response around critical regions, illustrating the effect of channel recalibration on the encoded features.

In the decoding stage, features reconstructed from the encoder without SE appear coarser and less localized across the nested fusion nodes. Features reconstructed from the SE-residual encoder maintain stronger responses around key anatomical regions and capture local structural details more effectively. These observations illustrate how recalibration inside the encoder propagates into subsequent multi-scale reconstruction.

\section{Discussion}
SRUG's reconstruction results support a direct encoder--decoder approach to paired MRI synthesis. The full-resolution reference and recalibrated feature pyramid supply spatial information that NMD repeatedly reuses, while MSS loss constrains the output across image scales. This formulation removes adversarial competition and iterative sampling; the present experiments establish synthesis quality, while matched measurements of optimization stability and inference cost remain necessary.

The BraTS 2023 ablations \cite{ref25} show that channel recalibration improves SSIM and LPIPS when coupled with NMD, while removing MSS loss reduces SSIM. These findings support the selected encoding--decoding configuration and structural supervision, without establishing a statistically significant interaction between components.

Independent evaluation on IXI \cite{ref27} supports applicability to another dataset and modality pair. Zero-shot transfer from BraTS 2023 \cite{ref25} to BraTS 2019 \cite{ref5} provides preliminary evidence of transferability, with mixed results across tasks and metrics. Broader scanner and protocol coverage, radiologist studies, and downstream diagnostic evaluations are needed to assess clinical utility.

\section{Limitations and future directions}
Several limitations remain. First, the present experiments quantify synthesis quality; matched analyses of optimization stability and inference cost, and direct comparisons with VAE-based models, remain for future work. Second, the current evaluation focuses on MRI data; performance on CT or PET has not been assessed. Third, 2.5D inputs \cite{ref37} provide only local inter-slice context and cannot fully model continuous 3D anatomy. Fourth, SRUG requires paired training data, and its deterministic output does not quantify conditional synthesis uncertainty. Extensions to unpaired settings and probabilistic prediction require further investigation. Finally, clinical applicability should be assessed through multi-center validation, radiologist studies, and downstream diagnostic tasks.

\section{Conclusion}
SRUG is a supervised standalone generator for paired MRI sequence synthesis. Its adapted residual encoder retains a full-resolution reference and recalibrated spatial features, which NMD repeatedly fuses to reconstruct the target. L1 and MSS loss jointly supervise this mapping, enabling synthesis in one forward evaluation. Experiments on BraTS 2023 and IXI demonstrate competitive reconstruction quality, while zero-shot testing on BraTS 2019 provides preliminary evidence of transferability. Further studies should quantify optimization stability, inference cost, and clinical utility.

\section*{Declaration of competing interest}\label{section:Declaration of competing interest}
The authors declare that they have no known competing financial interests or personal relationships that could have appeared to influence the work reported in this paper.

\section*{Acknowledgement}
This work is supported by Shanghai Municipal Natural Science Foundation (23ZR1425400), the Clinical Research Project of the
Shanghai Municipal Health Commission (20214Y0468), Zhongshan Hospital Management Science Foundation Project 2024ZSGL09 and Zhongshan Hospital Medical Humanities and Ideological Education Research Project 2024YXRWSZ-007.

\section*{Declaration of generative AI and AI-assisted technologies in the writing process.}
During the preparation of this work the authors used ChatGPT in order to improve language. After using this tool, the authors reviewed and edited the content as needed and take full responsibility for the content of the publication.

\bibliographystyle{elsarticle-num}
\bibliography{cas-refs}

\clearpage
\onecolumn
\appendix
\setcounter{figure}{0}
\setcounter{table}{0}
\setcounter{equation}{0}
\renewcommand{\thefigure}{A.\arabic{figure}}
\renewcommand{\thetable}{A.\arabic{table}}
\renewcommand{\theequation}{A.\arabic{equation}}
\section{Supplementary Error Heatmaps}
\captionsetup{hypcap=false}

\begin{center}
    \includegraphics[width=0.9\textwidth]{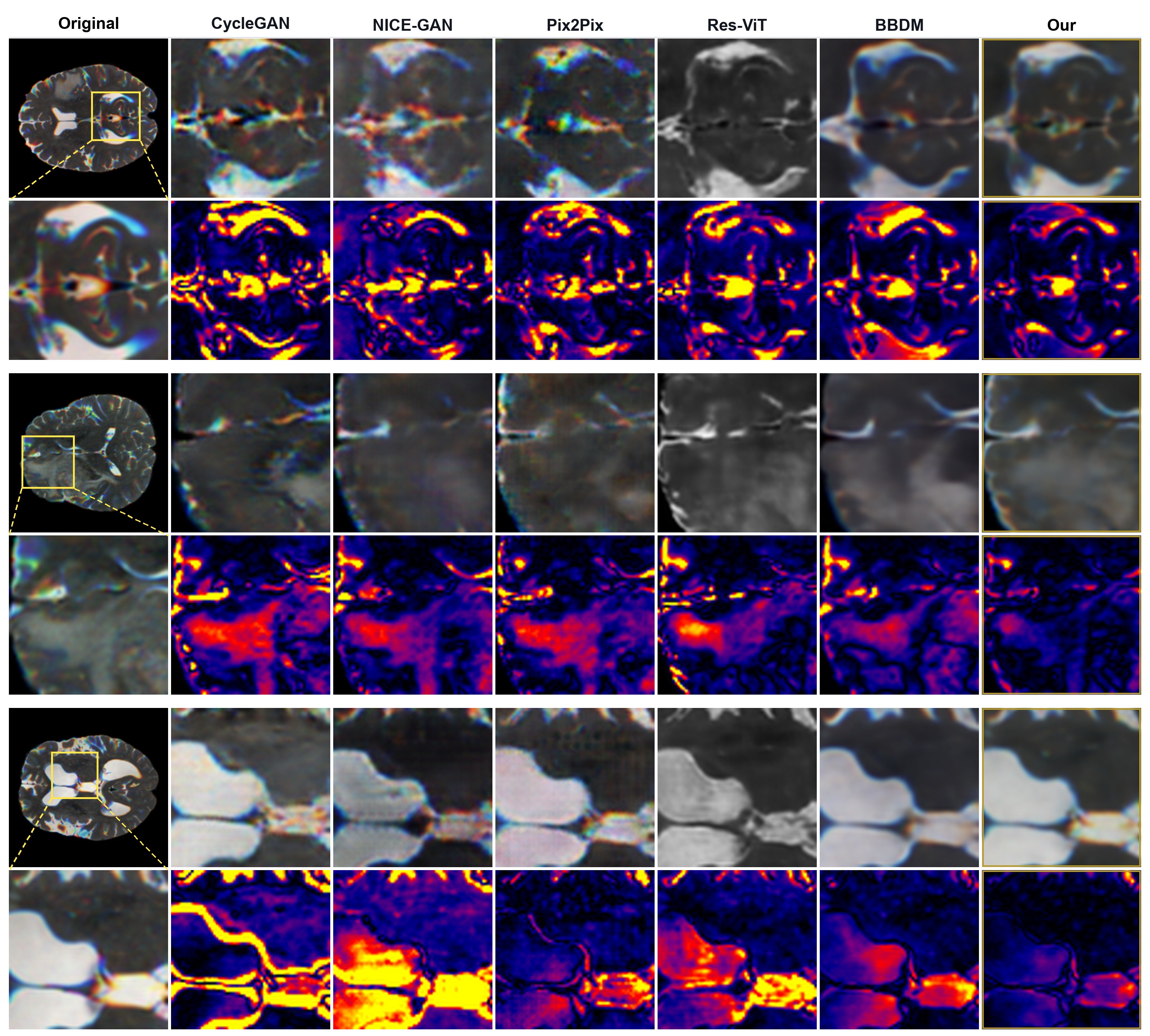}
    \captionof{figure}{Supplementary ROI error heatmaps; brighter regions indicate larger errors.}
    \label{fig:roi_heatmaps_part2}
\end{center}

\clearpage

\begin{center}
    \includegraphics[width=0.46\textwidth]{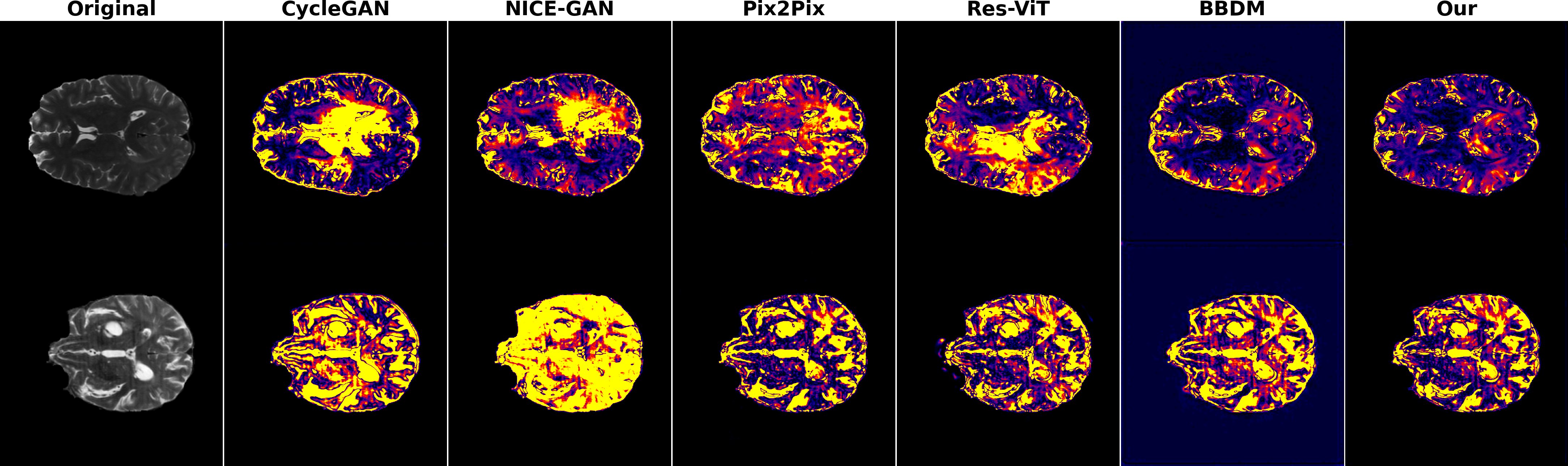}\\[-0.35em]
    \includegraphics[width=0.46\textwidth]{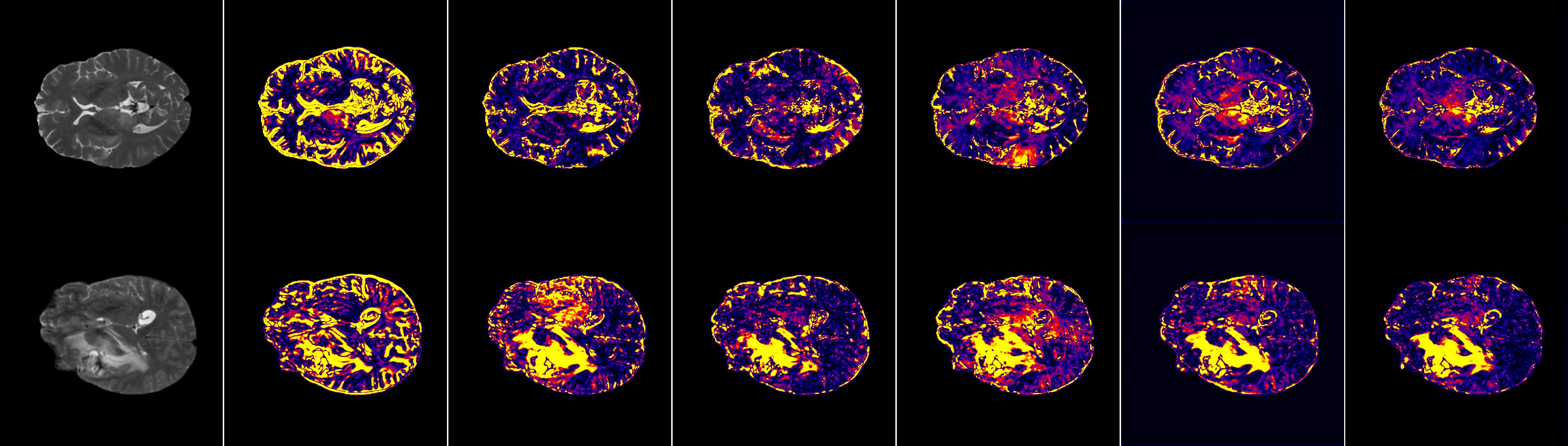}\\[-0.35em]
    \includegraphics[width=0.46\textwidth]{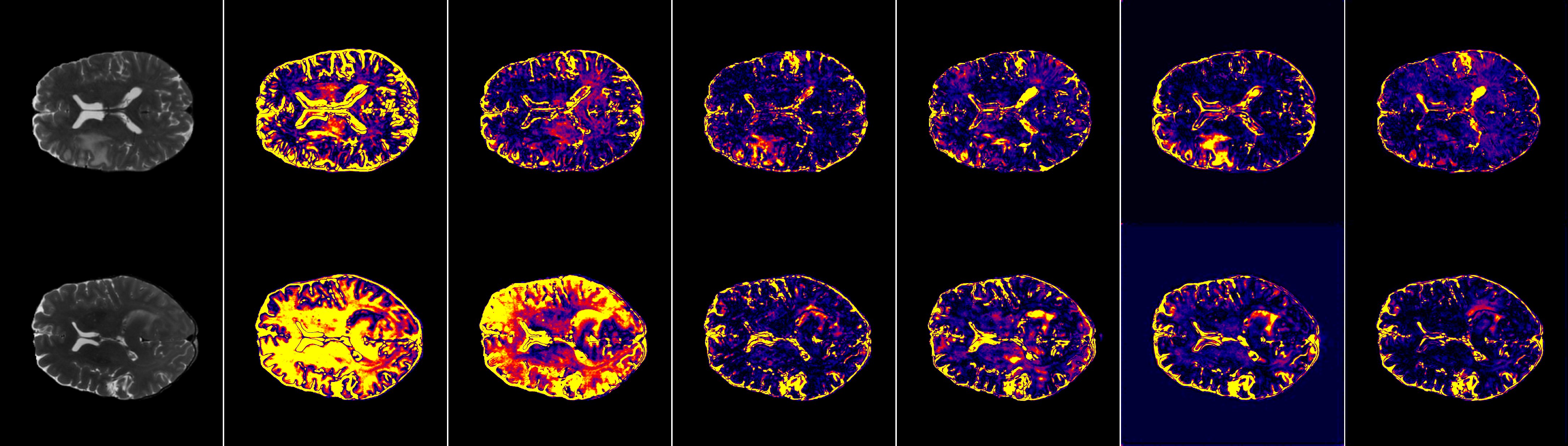}\\[-0.35em]
    \includegraphics[width=0.46\textwidth]{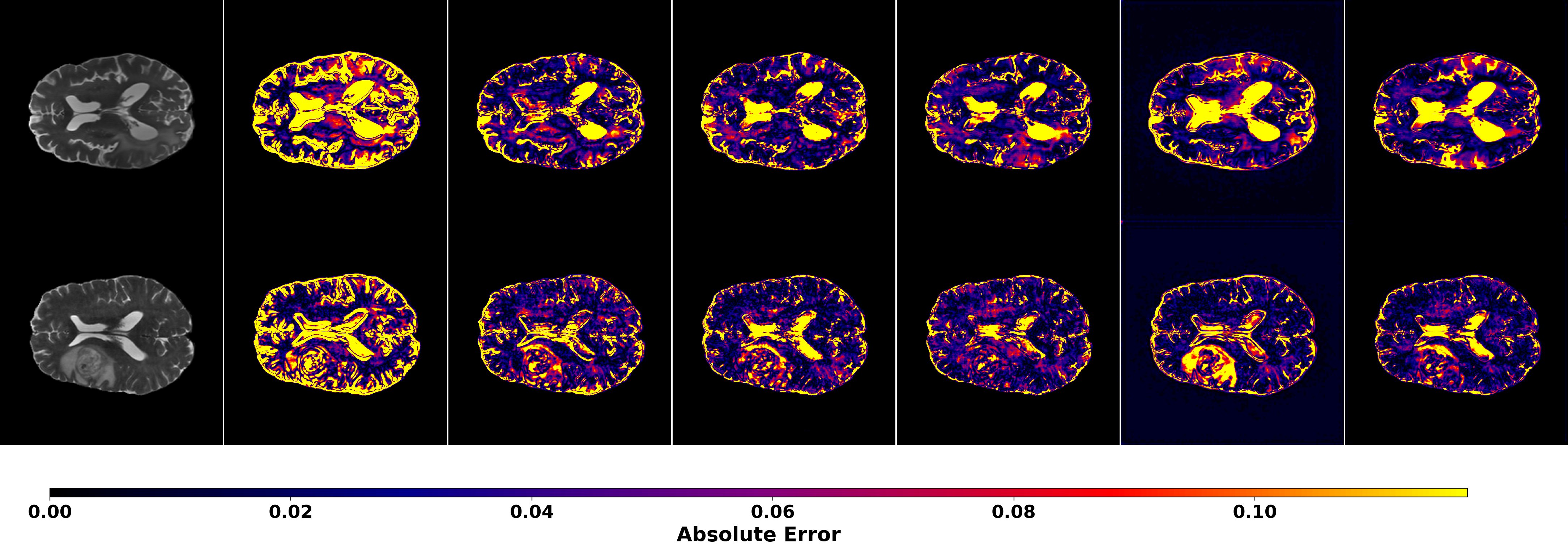}
    \captionof{figure}{Supplementary full-image error heatmaps; brighter regions indicate larger errors.}
    \label{fig:heatmaps_part2}

    \vspace{0.3em}
    \includegraphics[width=0.46\textwidth]{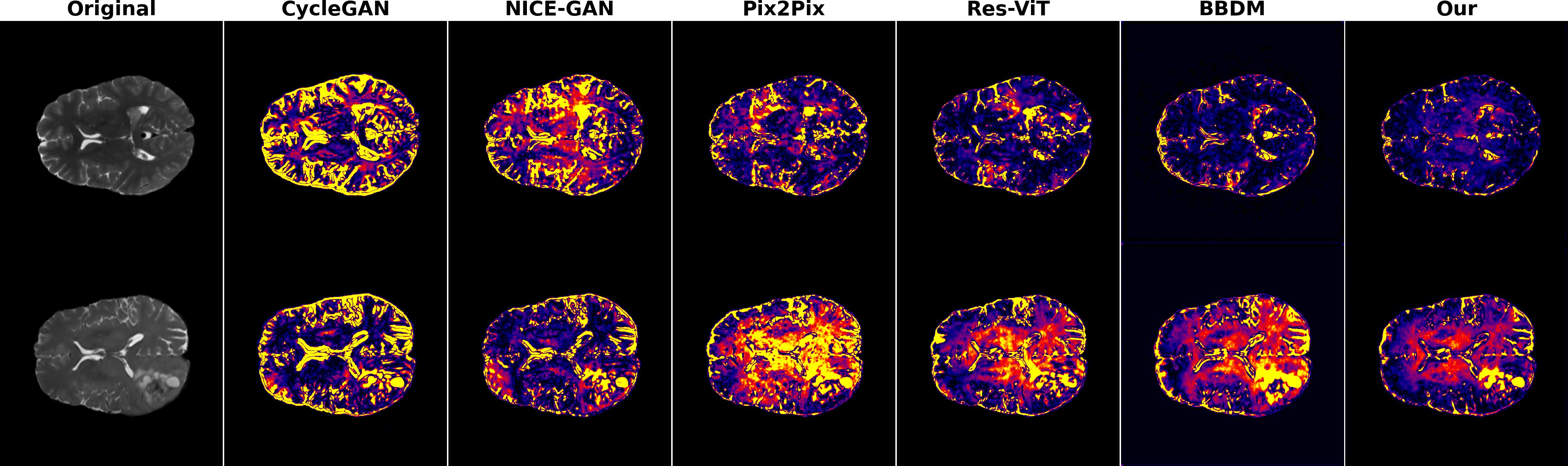}\\[-0.35em]
    \includegraphics[width=0.46\textwidth]{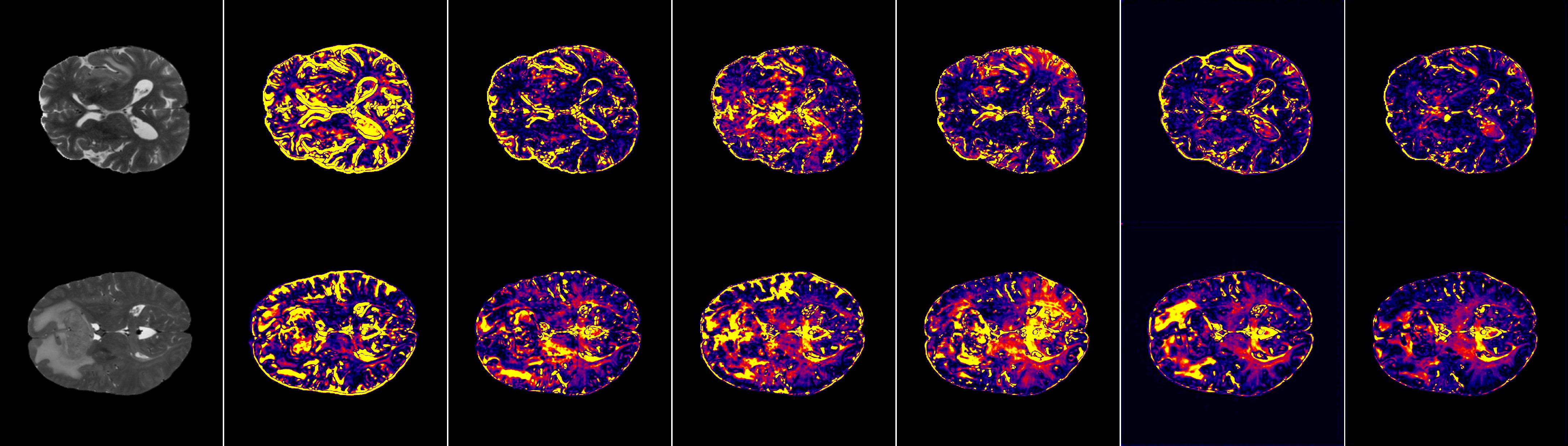}\\[-0.35em]
    \includegraphics[width=0.46\textwidth]{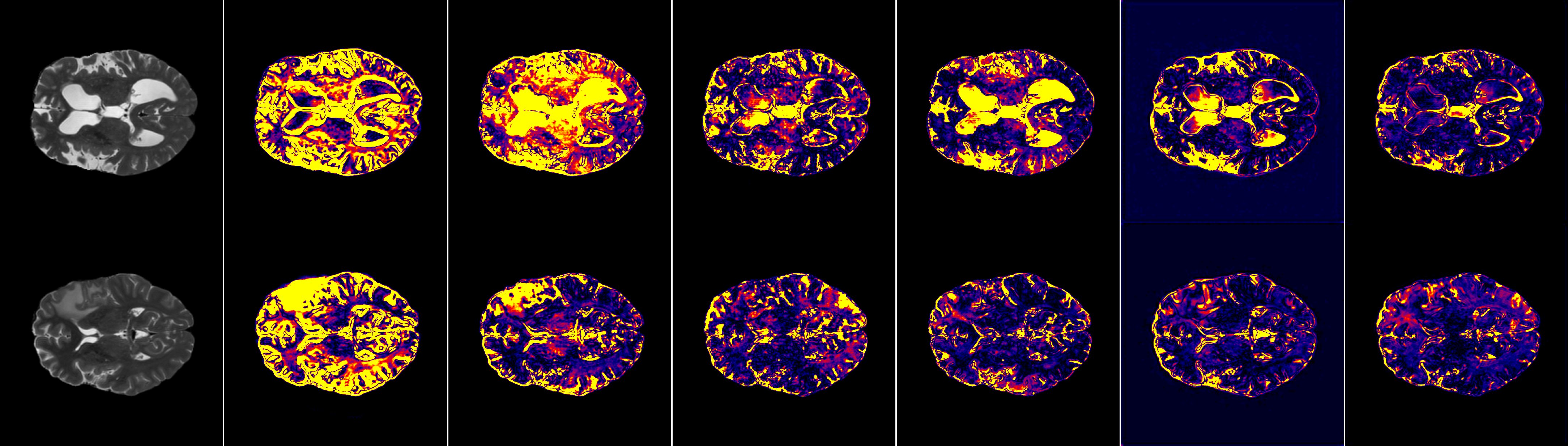}\\[-0.35em]
    \includegraphics[width=0.46\textwidth]{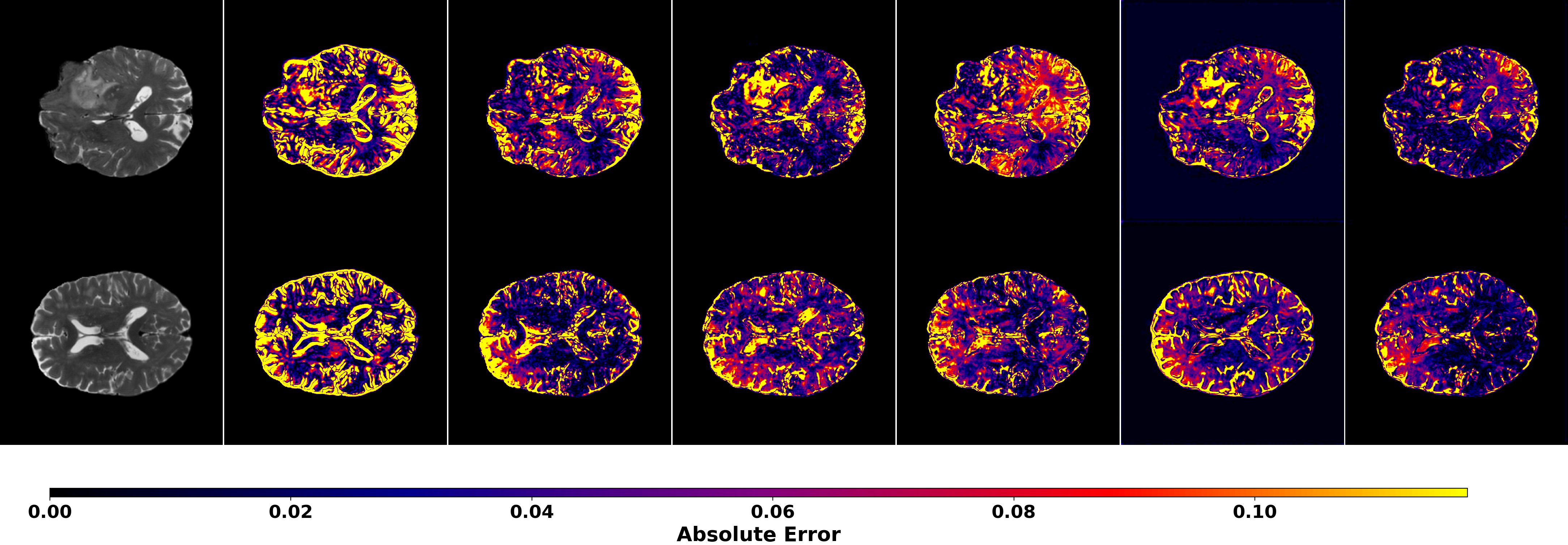}
    \captionof{figure}{Supplementary full-image error heatmaps; brighter regions indicate larger errors.}
    \label{fig:heatmaps_part3}
\end{center}

\clearpage

\begin{center}
    \includegraphics[width=0.46\textwidth]{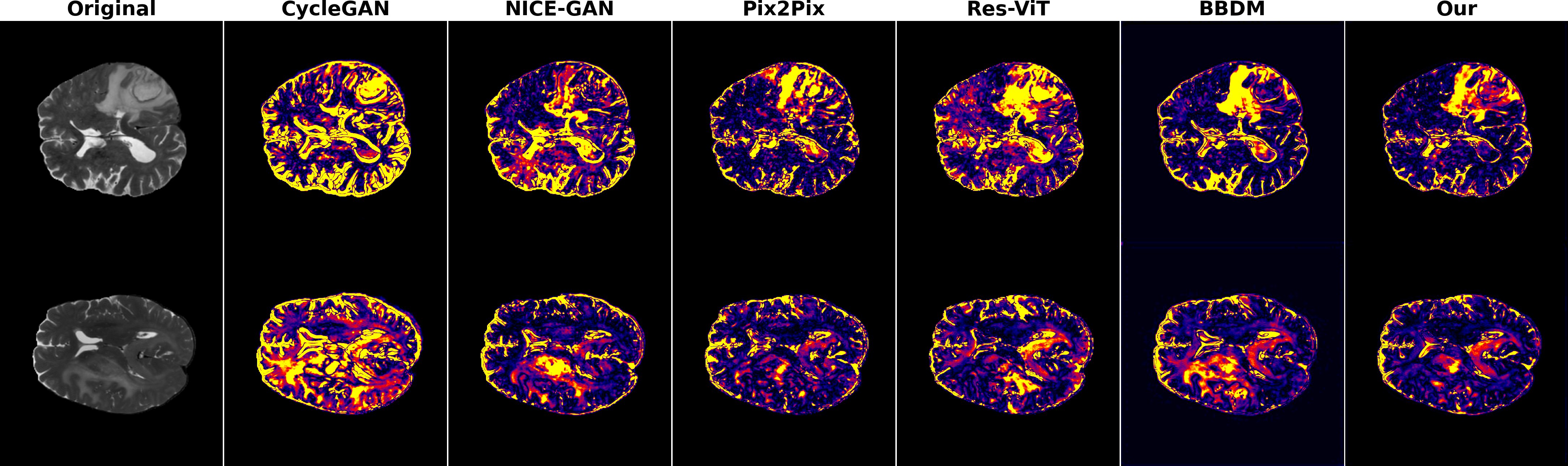}\\[-0.35em]
    \includegraphics[width=0.46\textwidth]{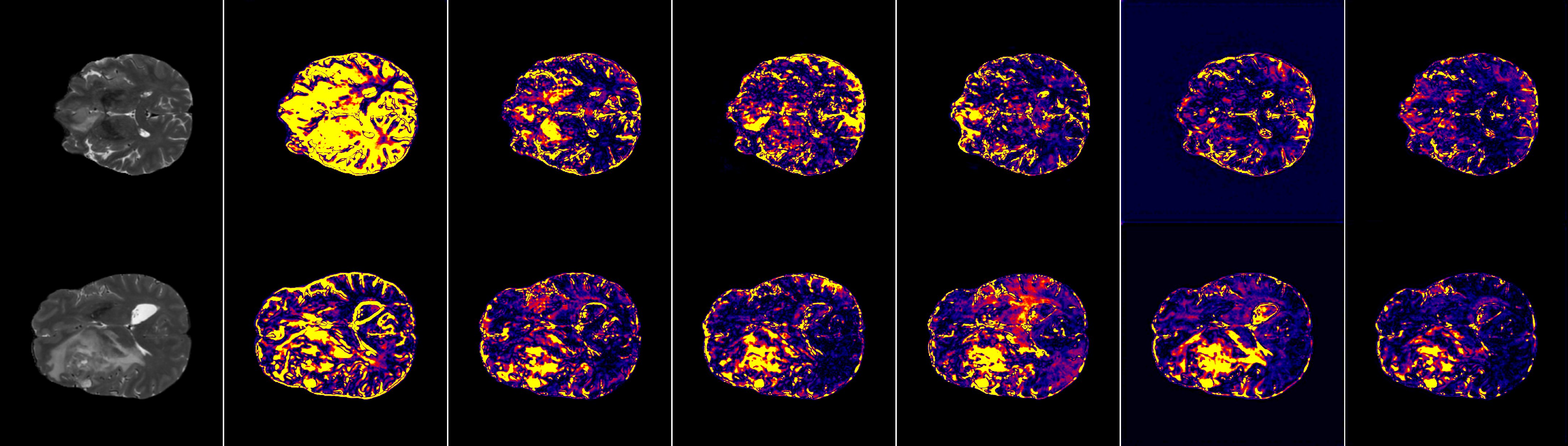}\\[-0.35em]
    \includegraphics[width=0.46\textwidth]{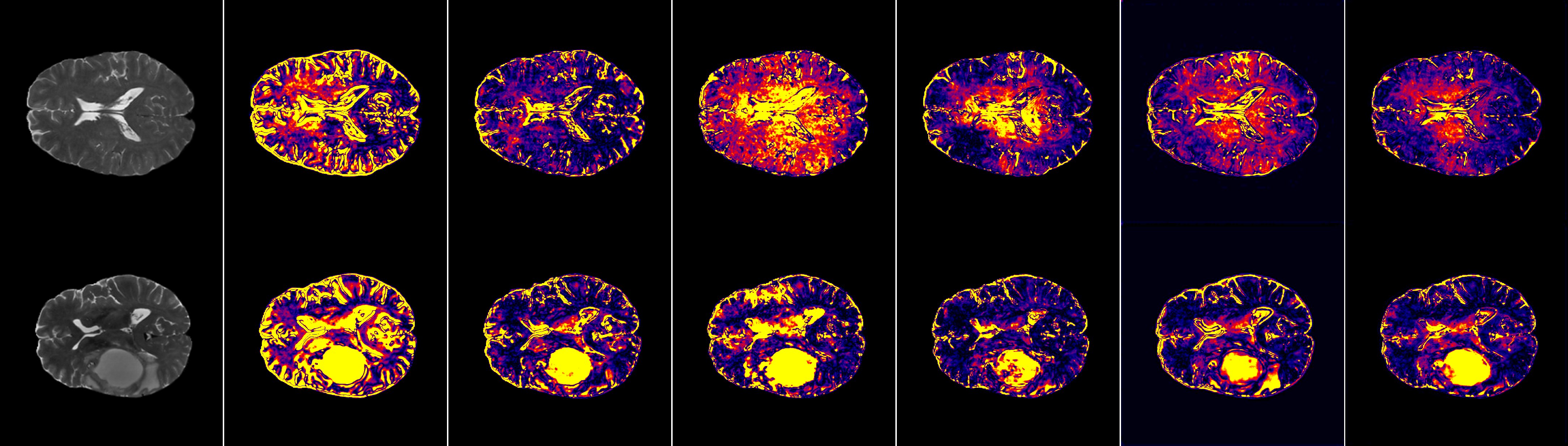}\\[-0.35em]
    \includegraphics[width=0.46\textwidth]{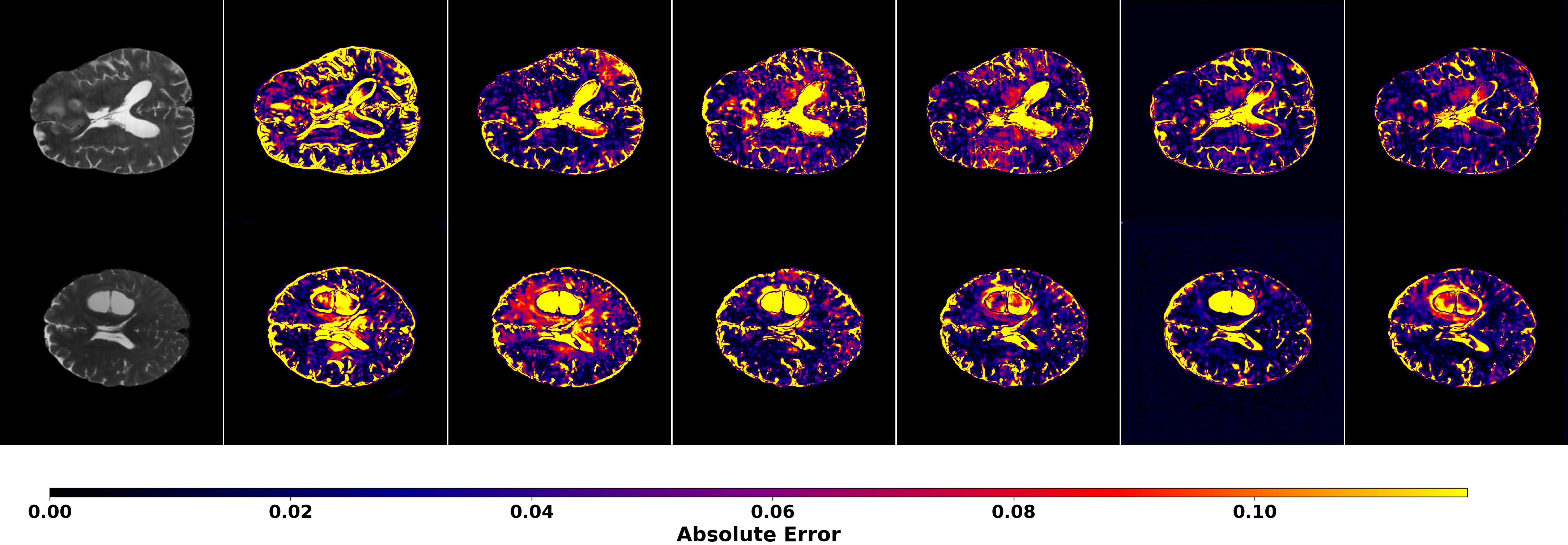}
    \captionof{figure}{Supplementary full-image error heatmaps; brighter regions indicate larger errors.}
    \label{fig:heatmaps_part4}

    \vspace{0.3em}
    \includegraphics[width=0.46\textwidth]{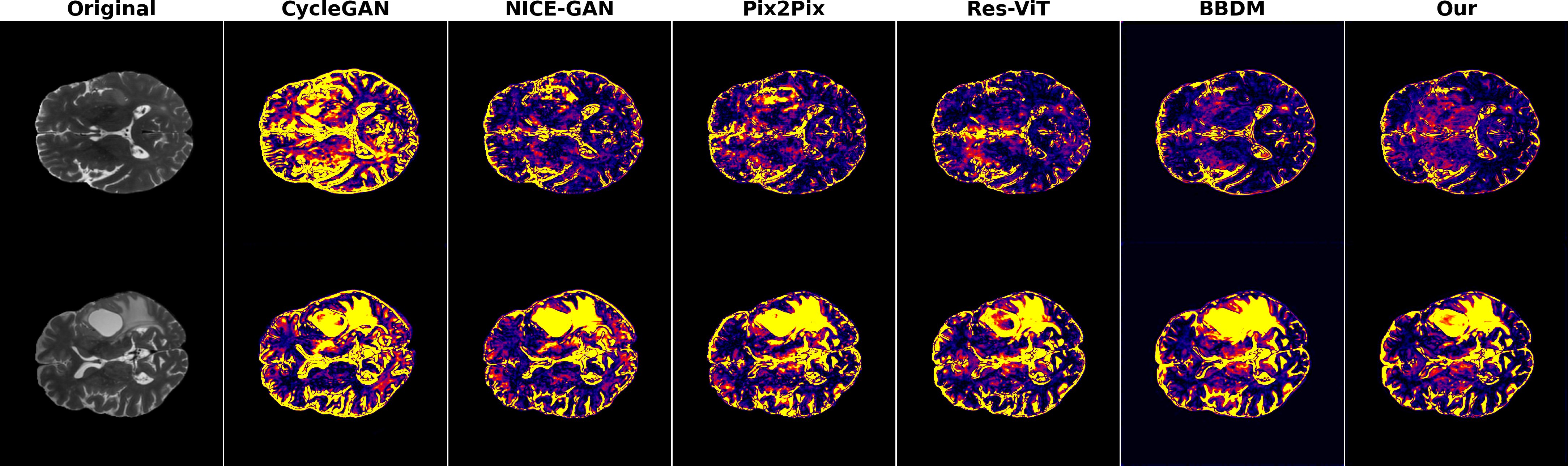}\\[-0.35em]
    \includegraphics[width=0.46\textwidth]{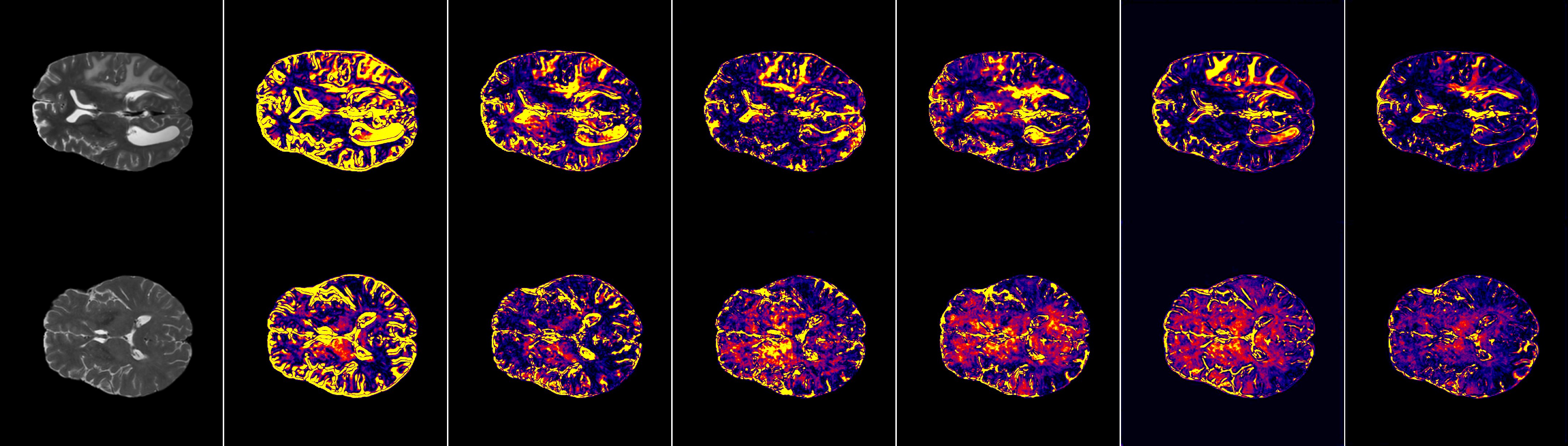}\\[-0.35em]
    \includegraphics[width=0.46\textwidth]{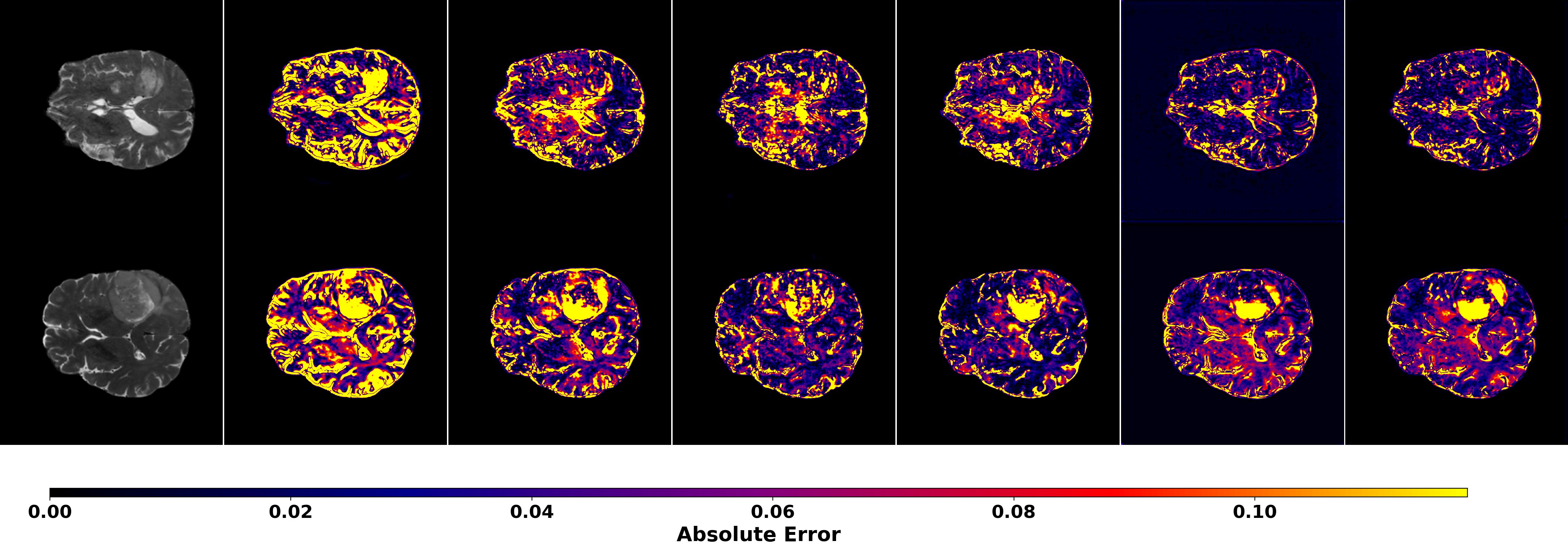}
    \captionof{figure}{Supplementary full-image error heatmaps; brighter regions indicate larger errors.}
    \label{fig:heatmaps_part5}
\end{center}

\end{document}